\documentclass[10pt,twocolumn,letterpaper]{article}
\ifdefined\pdfsuppressptexinfo\pdfsuppressptexinfo=-1\fi

\usepackage[pagenumbers]{cvpr} %

\newif\ifsuppl \newif\ifsupplonly
\suppltrue  \supplonlyfalse    %
\makeatletter
\newif\ifdroppages
\AddToHook{shipout/before}{\ifdroppages\DiscardShipoutBox\fi}
\long\def\suppl@iwrite#1#2#3{\begingroup#2\let\protect\@unexpandable@protect
  \edef\reserved@a{\immediate\write#1{#3}}\reserved@a\endgroup\if@nobreak\ifvmode\nobreak\fi\fi}
\newcommand{\startdroppingpages}{\droppagestrue
  \let\suppl@savedwrite\protected@write\let\protected@write\suppl@iwrite}
\newcommand{\stopdroppingpages}{\droppagesfalse\let\protected@write\suppl@savedwrite}
\makeatother

\usepackage{booktabs}

\newcommand{\paper}{HardSpotter\xspace}
\newsavebox{\fitbox}
\newcommand{\fitcolumn}[1]{\sbox{\fitbox}{#1}\ifdim\wd\fitbox>\columnwidth\resizebox{\columnwidth}{!}{\usebox{\fitbox}}\else\usebox{\fitbox}\fi}
\usepackage{tikz}
\usetikzlibrary{calc,fit,backgrounds,arrows.meta,positioning}

\DeclareMathOperator*{\argmax}{arg\,max}

\definecolor{cvprblue}{rgb}{0.21,0.49,0.74}
\usepackage[pagebackref,breaklinks,colorlinks,allcolors=cvprblue]{hyperref}

\def\confName{CVPR}
\def\confYear{2027}

\title{HotSpotter to \paper: Animal Re-Identification from Local Matches Above Chance}

\author{Fedor Zolotarev$^{1}$ \quad Aleksandr Algasov$^{2,4}$ \quad Ekaterina Nepovinnykh$^{2}$\\
Ankit Kumar Upadhyay$^{3}$ \quad Tuomas Eerola$^{2}$ \quad Pavel Zem\v{c}\'{i}k$^{4,2}$\\
Heikki K\"{a}lvi\"{a}inen$^{2,4}$ \quad Charles V. Stewart$^{3}$ \quad Enrico Di Minin$^{1}$\\[4pt]
$^{1}$University of Helsinki \quad
$^{2}$Lappeenranta-Lahti University of Technology LUT\\
$^{3}$Rensselaer Polytechnic Institute \quad
$^{4}$Brno University of Technology
}

\begin{document}
\ifsupplonly\startdroppingpages\fi
\maketitle
\begin{abstract}
Re-identifying individual animals from photographs underpins population estimation and the management of threatened species, but it is also quite difficult due to challenging imaging conditions, long term changes and subtle differences that distinguish individuals. A major obstacle is ground truth, since labeling a dataset is the very task that needs to be automated. Earlier methods, such as HotSpotter, required little prelabeled data, matching generic local features, with results easily visualized. More recent and more accurate end-to-end multispecies models, such as MiewID and MegaDescriptor, are trained on extensive labeled training data, relying on zero-shot transfer for new, unlabeled species, while sacrificing interpretability. We propose HardSpotter, a local feature method that closes the accuracy gap without labels and is inherently interpretable. Working with generic HesAffNet features and HardNet8 descriptors the key question is how to separate true inter-image matches from coincidences.  Using a gallery of descriptors for each species, HardSpotter does this by fitting a normalizing flow that maps the descriptors invertibly to an approximately-normal distribution; the chance level of a match then follows easily. Matching features are also compared using DINOv3 local tokens to ensure that they come from the same body part. The aggregate matching score of an image pair is a simple sum over its accepted matches. Despite requiring \textbf{no labeled individual ids}, HardSpotter achieves substantially   higher mean top-1 accuracy on WildlifeDatasets --- a suite of 29 animal re-identification datasets --- than all tested supervised and unsupervised methods, with the sole exception of the recent, supervised WildFusion, which HardSpotter trails by less than 1\%.
\end{abstract}

\section{Introduction}
\label{sec:intro}

\begin{figure*}[t]
  \centering
  \includegraphics[width=0.89\textwidth]{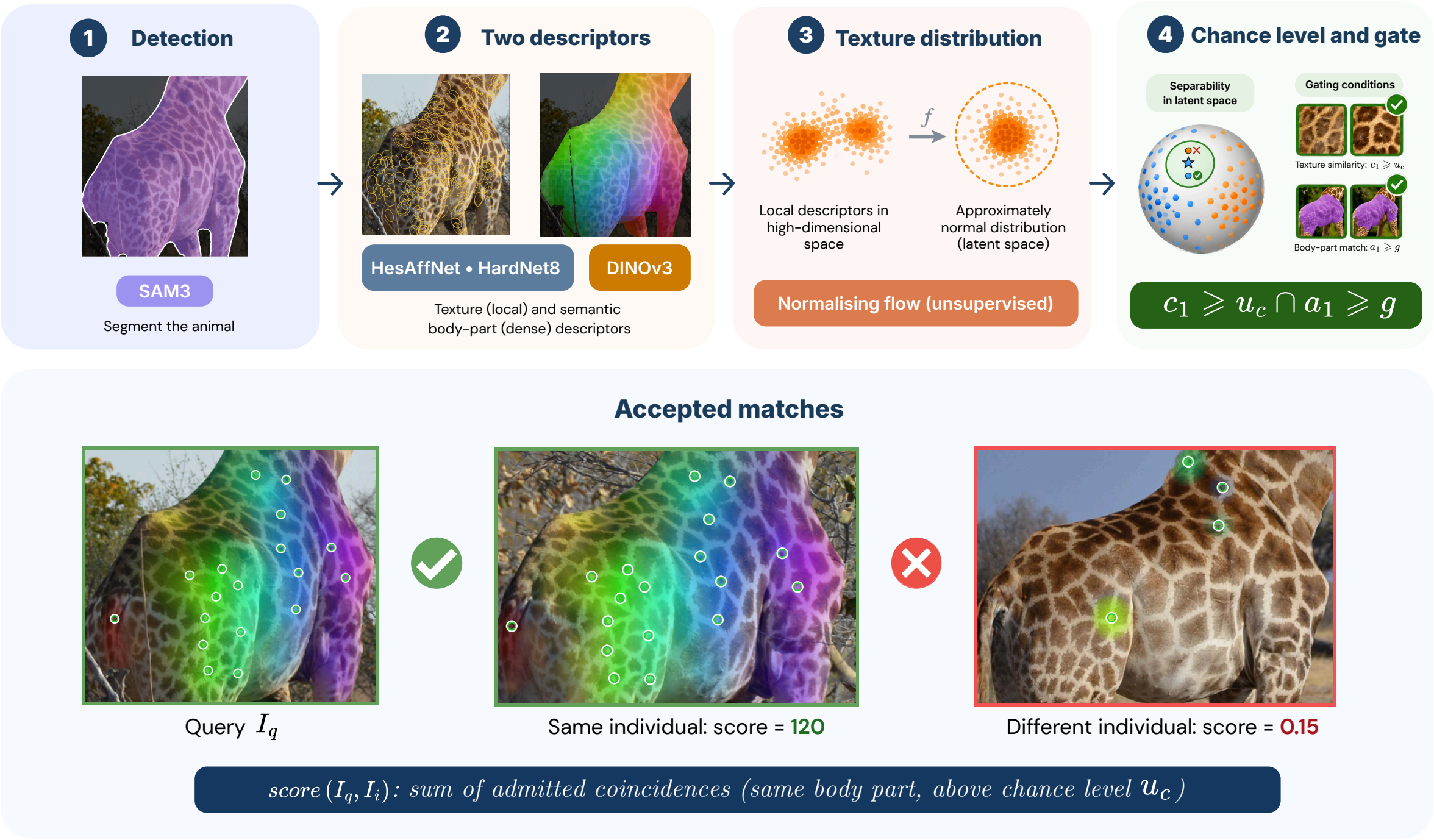}
  \caption{Overview of \paper. (1) SAM3 segments the animal. (2) Each keypoint gets a HardNet8 texture descriptor and a DINOv3 token for its place on the body. (3) A normalizing flow, fitted without labels, makes the texture descriptors approximately normal, so that the similarity of unrelated descriptors has a known distribution. (4) A match is accepted when its texture similarity exceeds the chance level $u_c$ and both keypoints lie on the same body part. Bottom: the score of an image pair sums its accepted matches, high for the same individual and near zero for a different one.}
  \label{fig:teaser}
\end{figure*}

Animal identification is the problem of deciding which images show the same individual animal~\cite{schneider2019past,vidal2021perspectives}. It lays the foundation for higher-level ecological and biological tasks, such as population estimates~\cite{parham2017animal}, individual tracking and the management of threatened species~\cite{tuia2022perspectives}.  As a starting point, it may be solved as a verification problem --- deciding whether or not a pair of images shows the same individual --- and this solution may be extended to the full id problem.

Animal id verification is challenging for a number of reasons:  (1) Some distinguishing features may appear in only one image due to viewpoint differences and occlusions.  (2) The body of the animal may bend and twist between the two photographs. (3) The animal itself may change throughout its life. (4)  Individuals of the same species are often similar in appearance, with only subtle features  distinguishing one individual from another. (5) Different species may have very different features that distinguish them --- the color pattern of the flank, the outline of the fluke or dorsal fin, or subtle markings of the face. (6) In many real-world settings, an individual of a species is only seen a few times and often only once~\cite{tuia2022perspectives,adam2024wildlifereid}.

The most difficult challenge is the seemingly mundane issue of assigning ground truth individual identity labels. Unless there is a source of metadata to narrow the search, this is a manual labeling process that requires a quadratic number of human expert verification decisions. This is both error prone --- especially for subtly-marked species --- and limits the size of datasets.

Early methods for animal id focused on matching local features and did not face this labeling problem as they had little dependence on training data.  The performance improvements seen by applying deep learning made this challenge important to address. One potential solution, explored in MegaDescriptor~\cite{cermak2024wildlifedatasets} and MiewID~\cite{otarashvili2024miewid}, is a form of transfer learning where a single model is trained on a number of species and applied to new species through either fine-tuning or zero-shot transfer. A second solution, explored in this paper, is to return to the application of generically-trained (zero shot) feature extraction and description, while avoiding the need for expensive-to-obtain training data.

We focus first on the question of ``how can we tell whether two local features show the same area of the same animal?'' This question is not new, and has traditionally been addressed heuristically using measures such as the ratio test~\cite{lowe2004distinctive} and LNBNN~\cite{mccann2012local}. Each of these roughly approximates how close two unrelated descriptors come by chance. The first and main insight of this paper is that this chance level can be computed in closed form from the empirical distribution of local feature descriptors taken from a set of unlabeled images of the species of interest. A normalizing flow~\cite{papamakarios2021normalizing} fitted to that distribution maps the descriptors into a space where they are approximately Gaussian, so the cosine similarity between unrelated descriptors becomes random with a known distribution. From that distribution we compute the similarity threshold at which two detections of different animals are expected to share about one chance match. Image pairs showing different animals then have about that many matches, whereas pairs showing the same animal have noticeably more, and the verification criterion is built on this contrast. 

Our second insight is obtained by revisiting the issue of geometric verification ~\cite{philbin2007object}, whereby a local feature match is eliminated if its corresponding image locations are inconsistent with an estimated global linear (often affine) transformation between the images. 
We replace this rough heuristic by developing a model of the empirical distribution of body features as represented by DINOv3 local tokens ~\cite{simeoni2025dinov3}, and use this to test whether two features are from the same body location.

We combine these two correspondence tests --- one of the distinctiveness of matching features and the other of their location consistency --- by developing statistically-derived threshold tests for each and simply summing, over the matches between two images that pass both tests, the margin by which each exceeds its threshold.  For matching a query image against a database (gallery) of images, we repeat this verification test by matching the query image against each gallery image and rank ordering the results. We call the method \paper\footnote{The code is publicly available at \href{https://gitlab.com/helics-lab/hardspotter}{https://gitlab.com/helics-lab/hardspotter}.}%
: it inherits the unsupervised local matching of HotSpotter~\cite{crall2013hotspotter} and matches HardNet8 descriptors, with the heuristics of both replaced by the statistically derived tests. The name joins the two.  

We evaluate on 29 animal re-identification datasets from WildlifeDatasets~\cite{cermak2024wildlifedatasets}. The baselines differ in what they learn from. MiewID~\cite{otarashvili2024miewid} and MegaDescriptor~\cite{cermak2024wildlifedatasets} are trained on animal identity labels, MegaDescriptor on these very galleries. WildFusion~\cite{cermak2024wildfusion} calibrates on the identity labels of half of the queries. The learned matchers LightGlue~\cite{lindenberger2023lightglue} and FlexGlue~\cite{lindenberger2026flexglue} are trained on correspondences from natural scenes and use no animal labels. HotSpotter~\cite{crall2013hotspotter} and ALFRE-ID~\cite{nepovinnykh2024alfreid}, like our method, use no identity labels at all.
Our method outperforms all methods in mean top-1 accuracy except WildFusion, which it trails by less than 1\%. 
On species with natural markings, the setting that local matching is made for, the proposed method is ahead of every other method on average. It trails behind MegaDescriptor, WildFusion, and FlexGlue only on face-based re-identification.

Our contributions are as follows:
\begin{enumerate}
  \item A re-identification method that runs on frozen, general-purpose models, extracting all training information from a gallery of images of a species of interest unsupervised, without any individual identity labels. The only model fit to the gallery is a normalizing flow over its own descriptors, learned without labels.

  \item  A location consistency test instead of geometric verification: DINOv3 local tokens accept a match only if the two keypoints lie at approximately the same place on the body.

  \item A closed-form chance level for local matching: a normalizing flow makes unrelated descriptors close to uniform directions. The threshold at which two different animals share about one match is derived in closed form and calibrated using an order statistic from the data. 

  \item An extension of the method to open-set re-identification with results showing 91\% accuracy in deciding when an individual is not in the gallery.
\end{enumerate}

\section{Related work}
\label{sec:related}

\paragraph{Animal re-identification.}
Original animal re-identification systems match local features between two images without supervision over the process: HotSpotter~\cite{crall2013hotspotter}, Wild-ID~\cite{bolger2012wildid}, StripeSpotter~\cite{lahiri2011stripespotter} and I$^3$S~\cite{vantienhoven2007i3s}.
Since 2013, most applications have shifted towards end-to-end models: MegaDescriptor~\cite{cermak2024wildlifedatasets} and MiewID~\cite{otarashvili2024miewid} are deep learning models trained on identity labels pooled across many animal species.
Among local-feature systems, ALFRE-ID~\cite{nepovinnykh2024alfreid}, a successor of NORPPA~\cite{nepovinnykh2024norppa} designed for ringed seals, extracts the pelage pattern with a trained segmentation network, aggregates local descriptors, HesAffNet and HardNet among them, into Fisher vectors~\cite{perronnin2010improving}, and re-ranks by the inliers of a homography fitted by RANSAC~\cite{nepovinnykh2024alfreid,immonen2023combining}.
WildFusion~\cite{cermak2024wildfusion} adds to MegaDescriptor the counts of learned-matcher matches above a confidence threshold, and fits the threshold and the fusion on labeled pairs.
The gain in accuracy came at the cost of labeled training data and of decisions that can no longer be inspected. We return to the first family and show that the accuracy need not be given up.

\paragraph{Local matching and chance.}
The ratio test~\cite{lowe2004distinctive} and nearest-neighbor classifiers~\cite{boiman2008defense} judge a match against a background distance read from the data. LNBNN~\cite{mccann2012local} reads it from the $(k+1)$-th nearest neighbor, an idea that our per-keypoint threshold takes up. Local scaling~\cite{zelnikmanor2005self} and hubness reduction~\cite{schnitzer2012local,conneau2018word} rescale similarities in the same spirit.
Retrieval pipelines score the descriptor pairs above a similarity threshold tuned for retrieval~\cite{tolias2013aggregate,tolias2014visual} or keep the matches consistent with one geometric transform~\cite{philbin2007object,barath2020magsac}, and re-rankers~\cite{suma2024ames,suma2026elvis} learn these rules from labeled image pairs.
ELViS~\cite{suma2026elvis} sums thresholded best matches, as we do, but learns the threshold from labeled pairs.
A threshold derived from a background model is the idea of a-contrario detection~\cite{desolneux2000meaningful,moisan2004probabilistic,rabin2009statistical,sabater2012meaningful}, which declares an event meaningful when its expected number by chance is below one.
We keep its convention and obtain the background in closed form.

\paragraph{Learned matchers.}
SuperGlue~\cite{sarlin2020superglue} replaced the nearest-neighbor rule with a network that matches the keypoints of two images jointly, by attention within and across the images.
LightGlue~\cite{lindenberger2023lightglue} revisits this design and adapts its computation to the difficulty of the pair.
LoFTR~\cite{sun2021loftr} and RoMa~\cite{edstedt2024roma} match dense feature grids without a detector.
FlexGlue~\cite{lindenberger2026flexglue} matches the keypoints of one image against dense features of the other, computed from a frozen backbone trained with DINOv3, and learns from labeled pairs~\cite{cai2023doppelgangers} to reject pairs that look alike but are different.
OmniGlue~\cite{jiang2024omniglue} and Cadar et al.~\cite{cadar2024leveraging} use DINOv2 features to guide which keypoints a keypoint attends to or to weight its candidate matches, with geometric verification still downstream.
WildFusion~\cite{cermak2024wildfusion} uses these matchers on animals without fine-tuning, and the calibrated fusion of its LightGlue and LoFTR match counts is by itself ahead of MegaDescriptor on 11 of 17 datasets.
Their rules for when a match counts are learned from labeled correspondences of rigid scenes. The same decision is what we derive instead of learn.

\section{Proposed Method}
\label{sec:method}

\subsection{Preprocessing and local features}
\label{sec:preprocessing}
\label{sec:representation}

In order to ensure that only relevant features are used for re-identification, the first stage is data preprocessing, which consists of segmenting the target animal from the photograph. We solve this by using an open-vocabulary foundation segmentation model SAM3~\cite{carion2025sam3}, which detects animals when prompted with species names.
Many datasets in our experiments are published already processed or with segmentation masks, and for these we skip our own preprocessing and use the datasets as they are. More detailed information about datasets and their preprocessing is given in the supplementary material.

The method uses local features of two types: semantic and texture descriptors. For texture, features are detected by HesAffNet~\cite{mikolajczyk2004scale,mishkin2018repeatability} and described by HardNet8~\cite{mishchuk2017working,pultar2020improving}. HesAffNet is designed to detect affine and scale invariant features. 
The method uses a pretrained HardNet8 model, trained once for patch matching on photographs of landmarks and outdoor scenes, to produce the texture descriptor, a unit vector in $\mathbb{R}^{128}$, $t_i \in \mathbb{S}^{127}$. Semantic information is described by DINOv3 ViT-L/16~\cite{simeoni2025dinov3}, trained with self-supervision on web images. Patch tokens from the last layer are bilinearly interpolated at texture keypoints and normalized to unit length, resulting in $s_i \in \mathbb{S}^{1023}$. Neither model has seen animal identities.

\subsection{Gaussianizing the texture descriptors}
\label{sec:flow}

To decide whether two descriptors are more similar than chance, the method must know how similar the descriptors of two unrelated patches come by chance. On the raw descriptors this distribution is out of reach, because textures cluster, a stripe edge resembles every other stripe edge. We therefore apply a normalizing flow~\cite{papamakarios2021normalizing} to the texture descriptors, making their distribution approximately Gaussian. Under a Gaussian every direction is equally likely, the similarity of two unrelated descriptors has a known distribution, and the matching rules follow in closed form.
The flow works in Euclidean space and is not directly suited to a density on the sphere~\cite{gemici2016normalizing}, where the unit descriptors $t_i \in \mathbb{S}^{127}$ lie, so they are first mapped to $x \in \mathbb{R}^{127}$ by the logarithmic map (details in the supplementary material).

\paragraph{Flow.}
A normalizing flow~\cite{rezende2015variational,papamakarios2021normalizing} is an invertible map $f\colon \mathbb{R}^D \to \mathbb{R}^D$ that is trained via maximum likelihood on the descriptors of the gallery, without labels, to transform the descriptors $x$ into standard normal variables $z=f(x)$, so that the descriptors have the density $p(x) = \mathcal{N}(f(x); 0, I)\,\lvert \det J_f(x) \rvert$.

The direction of a standard Gaussian vector is uniformly distributed on the sphere, as used in \cref{sec:null}.
Before two keypoints are compared, each latent is centered on the mean latent of its detection. This is done to remove shared factors, such as lighting and camera. Then the keypoints are normalized to unit length, and $c(i, j)$ denotes the cosine of two such directions.

\subsection{Chance-level threshold}
\label{sec:null}

As is standard for unit vectors, two descriptors are compared by the cosine between them: the smaller the angle between them, the closer the cosine is to 1 and the more similar they are. If two features describe the same area, the angle between them should be smaller than between a pair of unrelated features. The directions of standard normal vectors are uniformly distributed on $\mathbb{S}^{D-1}$, so for two keypoints of different animals, $\tfrac12 (c + 1)$ follows a Beta distribution with both parameters equal to $(D-1)/2$~\cite{cai2013distributions}, so 
\begin{equation}
  S(u) = \Pr(c \ge u) = I_{(1-u)/2}\!\Big(\tfrac{D-1}{2}, \tfrac{D-1}{2}\Big),
  \label{eq:S}
\end{equation}
where $I$ is the regularized incomplete beta function.
The best of the $n_B$ cosines of a query keypoint in a gallery detection $B$ exceeds $u$ with probability $1 - (1 - S(u))^{n_B}$.
Following a-contrario detection~\cite{desolneux2000meaningful}, we set a single threshold $u_c$ for the whole gallery, fixed before any query is seen.
Its level is chosen so that, in a pair of unrelated detections, the expected number of query keypoints exceeding it by chance is one: chance then accounts for about one match in any pair of different animals, and every match beyond that is evidence that the two detections show the same individual.
Setting the expectation to one is the a-contrario convention, which replaces a free parameter by a stated false-alarm level.
With $\bar N$ the mean number of keypoints of a query detection and $\mathcal{G}$ the gallery, $u_c$ is the root of
\begin{equation}
  \frac{1}{|\mathcal{G}|} \sum_{B \in \mathcal{G}} \Big[ 1 - \big(1 - S(u_c)\big)^{n_B} \Big] = \frac{1}{\bar N}.
  \label{eq:uc}
\end{equation}
$u_c$ depends only on $D$, $\bar N$ and the keypoint counts of the gallery, and has no free parameter.

\paragraph{Per-feature calibration.}

The flow minimizes the Kullback–Leibler divergence between the resulting distribution and the standard normal, but does not guarantee a perfect match. The resulting distribution can deviate from normal, especially for highly redundant features. For example, a generic patch or plain fur has more look-alikes in the gallery than the closed form of \cref{eq:S} expects, and a rare marking fewer. One global threshold cannot account for this residual, so we calibrate the threshold of each query keypoint by comparing how many accidental look-alikes it has in the gallery with what the closed form expects. Accidental look-alikes can only be counted among detections of other animals, but the gallery most likely contains several detections of the query's own animal. The first step is therefore to set aside, without identity labels, the gallery detections that likely show the query's own animal. The idea is simple: apply the same reasoning from \cref{eq:uc}, but one step higher --- at the level of whole detections instead of individual keypoints. 

\Cref{eq:uc} sets $u_c$ so that about one keypoint of the query is accepted by accident in a detection of another animal.
One level higher, instead of individual keypoints the units are the gallery detections. $N_{AB}$ is the number of keypoints of $A$ accepted by $u_c$ in $B$.
In a detection of another animal, each keypoint of $A$ is accepted only rarely, with the probability $p_B$ from \cref{eq:uc}, so $N_{AB}$ is a count of rare events and follows a Poisson distribution.
Its mean is the closed form's prediction for the pair, $n_A\, p_B$, the $n_A$ keypoints of $A$ times the probability of each being accepted, corrected by $\bar r_{\mathrm{all}}$, the ratio of observed to expected accepted matches over all pairs of gallery detections: $\mu_{AB} = \bar r_{\mathrm{all}}\, n_A\, p_B$.
$B$ is taken to show the query's animal when a count as high as $N_{AB}$ would arise by accident in fewer than one of the $\lvert\mathcal{G}\rvert$ gallery detections:
\begin{equation}
  \Pr\big(\mathrm{Poisson}(\mu_{AB}) \ge N_{AB}\big) \le 1/\lvert\mathcal{G}\rvert.
  \label{eq:flag}
\end{equation}
These same-animal candidates are left out of the calibration, but their matches are still scored.
The same step, taken once more, decides whether the query's individual is in the gallery at all (\cref{sec:openset}).

For a keypoint $i$ of the query, its best match in a gallery detection $B$ is its most similar keypoint there, with cosine $c_1(i, B)$.
Over the detections that are not same-animal candidates, sort these best-match cosines as $c_{i,(1)} \ge c_{i,(2)} \ge \dots$.
At the level $c_{i,(k+1)}$, exactly $k+1$ detections hold a match of $i$ at or above it, while the closed form expects $L_i = \sum_B \big[1 - (1 - S(c_{i,(k+1)}))^{n_B}\big]$ of them.
The ratio $r_i = (k+1)/L_i$ is how many times more look-alikes $i$ has than the closed form expects, and with $\bar r$ the same ratio measured over pairs of gallery detections, same-animal candidates excluded, the threshold of $i$ is the root $u_i$ of
\begin{equation}
  S(u_i) = S(u_c)\, \frac{\bar r}{r_i},
  \label{eq:ui}
\end{equation}
so that every query keypoint collects accidental matches at the gallery's average rate.
Reading the background from the $(k+1)$-th nearest neighbor is the idea of LNBNN~\cite{mccann2012local}, whose $k = 10$ we keep, but there the depth is fixed, and on a gallery with many photographs of the same subject, such as a landmark benchmark, the $(k+1)$-th neighbor is itself a genuine match. Here the depth grows with the number of same-animal candidates, providing a more reliable estimate of the background level.

\subsection{Location consistency}
\label{sec:gate}

The same texture recurs along a body and across animals, so a spot pattern on the shoulder of one leopard can resemble one on the thigh of another.
A match is therefore admitted only if the two keypoints sit at the same place on the body, as read from their location tokens.

Since DINO is a general-purpose vision foundation model, most of its feature space encodes distinctions that never occur within one animal dataset.
Along a direction in which the gallery's tokens barely vary, every pair of keypoints looks alike, so that direction cannot tell two places on the body apart.
We therefore compare two tokens only along the directions in which the gallery's tokens do vary, their number read from the spectrum of the tokens' covariance, together with their mean direction, which carries the shared shape of the body.
The cosine of two tokens in this subspace is their location agreement $a(i, j)$.

The test needs a threshold on $a$ that separates genuine partners from look-alikes, and the gallery provides examples of both without labels.
A query keypoint shows one particular spot on the animal's surface, and that spot appears in a photograph at most once, however often its texture repeats elsewhere on the body.
So for a query keypoint $i$ and a gallery detection $B$ from another photograph, the keypoint $j^{\ast}$ of $B$ with the largest texture similarity to $i$ may be its genuine partner, while the one with the second-largest, $j^{\ast\ast}$, is almost always a look-alike.
Over the gallery against itself, the histogram of $a_2 = a(i, j^{\ast\ast})$ therefore shows how look-alikes agree in location, and the histogram of $a_1 = a(i, j^{\ast})$ mixes look-alikes with genuine partners, which agree more.
The threshold is placed where the two histograms cross: with $S_1(a) = \Pr(a_1 \ge a)$ and $S_2(a) = \Pr(a_2 \ge a)$,
\begin{equation}
  g = \argmax_{a \in [-1, 1]} \big[ S_1(a) - S_2(a) \big],
  \label{eq:gate}
\end{equation}
the value that maximizes the share of genuine partners it admits minus the share of look-alikes it admits.

\subsection{Score}
\label{sec:score}

The score of a gallery detection $B$ for the query $A$ adds, over the keypoints of $A$, the margin by which the best texture match exceeds the keypoint's threshold, for the matches that pass the location consistency test:
\begin{equation}
  \mathrm{score}(A, B) = \sum_{i \in A} \big( c_1(i, B) - u_i \big)_+ \;
  \mathbf{1}\big[ a_1(i, B) \ge g \big],
  \label{eq:score}
\end{equation}
where $B$ ranges over the gallery detections from other photographs than $A$.
Every term of the score is a pair of keypoints that can be drawn on the two photographs (\cref{fig:qualitative}).

\subsection{Open set}
\label{sec:openset}

The ranking always names a best gallery detection, even when the query's individual is not in the gallery. When dealing with an open set setting, where a query individual might not be present in the gallery, the intuitive way to decide is a threshold on the score, which decides whether to accept a match or not. Since every quantity in our score has a statistical meaning, that threshold can be estimated for each query, without supervision or manual tuning. Animals of the same species look alike, so even detections of other animals give the query some score. In a gallery of a reasonable size, most detections show other animals. Therefore, for a query of a known individual, its scores to the matching detections would rise above the scores to the other individuals, which determine the level of accidental scores.

This is the calibration of \cref{eq:ui} with scores in place of cosines. Scores have no closed form, so their spread is estimated from the query itself. As for $c_{i,(k+1)}$, the level $b_A$ is the $(k+1)$-th highest score outside the same-animal candidates, and $m_A$ detections score above it.
We assume that the top scores of other animals decay exponentially, as the top of a large sample does when its distribution has no heavy tail.
Plotted against the logarithm of their rank, such scores lie on a straight line, and its slope $\hat\sigma_A$ is estimated from the scores just below $b_A$, where only other animals lie.
The query is accepted when its best score $y_{(1)}$ lies so far above this line that other animals alone would reach it with probability at most $1/\lvert\mathcal{G}\rvert$, as in \cref{eq:flag}. That probability has a closed form, given in the supplementary material.

\section{Experiments}
\label{sec:experiments}

\subsection{Setup}
\label{sec:setup}

We evaluate on 29 datasets of the WildlifeDatasets toolkit~\cite{cermak2024wildlifedatasets}, each under its published closed-set split, with the training images as the gallery and the test images as the queries, and report top-1 accuracy.
MiewIDv3~\cite{otarashvili2024miewid} and MegaDescriptor~\cite{cermak2024wildlifedatasets} are trained on identity labels.
WildFusion~\cite{cermak2024wildfusion} calibrates its fusion on the identity labels of half of the queries, so we calibrate on one half, score the other and then swap the halves, and report only the scored halves.
ALFRE-ID~\cite{nepovinnykh2024alfreid}, HotSpotter~\cite{crall2013hotspotter}, which scores RootSIFT features by LNBNN with spatial re-ranking, and the learned matchers use no identity labels.
LightGlue~\cite{lindenberger2023lightglue} on DISK features~\cite{tyszkiewicz2020disk} and FlexGlue~\cite{lindenberger2026flexglue} on ALIKED keypoints~\cite{zhao2023aliked} score a pair by its number of matches, on the same images as our method with 2,048 keypoints per image.
To save computation, $g$, $\bar r_{\mathrm{all}}$ and $\bar r$ are estimated from the gallery against a random sample of 128 of its photographs. Each is a single number pooled over all keypoints of the gallery, so the sample adds far less variation than the fit of the flow, and on SealID the whole gallery gives the same top-1 accuracy at four times the cost.
Our results are the mean over five runs with different seeds, each fitting its own flow and drawing its own sample, and their standard deviation is at most 0.9 points on any dataset.

\subsection{Results on WildlifeDatasets}
\label{sec:main}

\begin{table*}[t]
  \centering
  \caption{Top-1 accuracy (\%) on WildlifeDatasets, grouped by the trait that carries identity~\cite{kuhl2013animal,vidal2021perspectives}. Best per row in bold. $^\dagger$MegaDescriptor's result, as WildFusion cannot be calibrated on this dataset.}
  \label{tab:main}
  \small
  \setlength{\tabcolsep}{4pt}
    \begin{tabular}{lccc@{\hspace{4\tabcolsep}}ccccc}
  \toprule
   & \multicolumn{3}{c}{Identity labels} & \multicolumn{5}{c}{No identity labels} \\
  \cmidrule(lr){2-4}\cmidrule(l){5-9}
  Dataset & MiewIDv3 & MegaDescriptor & WildFusion & ALFRE-ID & HotSpotter & LightGlue & FlexGlue & \textbf{\paper} \\
  \midrule
  \multicolumn{9}{l}{\emph{Pattern $\bullet$}} \\
  AAUZebraFish & 98.6\% & \textbf{99.9\%} & \textbf{99.9\%}$^\dagger$ & 83.5\% & 92.5\% & 99.3\% & 98.0\% & 99.5\% \\
  AerialCattle2017 & \textbf{100.0\%} & \textbf{100.0\%} & \textbf{100.0\%} & \textbf{100.0\%} & \textbf{100.0\%} & \textbf{100.0\%} & \textbf{100.0\%} & \textbf{100.0\%} \\
  ATRW & 98.6\% & 98.8\% & \textbf{99.1\%} & 98.6\% & 97.7\% & 96.5\% & 94.5\% & 98.7\% \\
  BirdIndividualID & 87.6\% & 97.8\% & \textbf{97.9\%} & 83.4\% & 87.6\% & 95.8\% & 96.1\% & 95.1\% \\
  Cows2021 & 98.8\% & 99.8\% & \textbf{100.0\%} & 98.6\% & 91.0\% & 96.6\% & 98.8\% & 97.0\% \\
  FriesianCattle2015 & \textbf{100.0\%} & \textbf{100.0\%} & \textbf{100.0\%} & 90.2\% & \textbf{100.0\%} & \textbf{100.0\%} & \textbf{100.0\%} & \textbf{100.0\%} \\
  FriesianCattle2017 & 89.4\% & 96.5\% & 90.4\% & 84.3\% & 79.8\% & 83.8\% & 66.7\% & \textbf{98.0\%} \\
  Giraffes & 98.5\% & 90.7\% & 98.9\% & 90.7\% & 98.9\% & \textbf{99.3\%} & 98.9\% & 98.9\% \\
  GiraffeZebraID & 80.6\% & 83.4\% & 94.7\% & 68.9\% & 90.2\% & 83.0\% & 74.5\% & \textbf{97.1\%} \\
  HyenaID2022 & 87.0\% & 78.6\% & \textbf{89.0\%} & 49.7\% & 67.9\% & 81.7\% & 79.4\% & 88.3\% \\
  LeopardID2022 & 79.6\% & 75.8\% & 85.6\% & 64.5\% & 87.2\% & 85.3\% & 78.6\% & \textbf{88.4\%} \\
  NyalaData & \textbf{81.1\%} & 38.3\% & 43.9\% & 11.9\% & 22.4\% & 14.7\% & 7.7\% & 54.2\% \\
  OpenCows2020 & 98.5\% & \textbf{100.0\%} & \textbf{100.0\%} & 98.6\% & 88.3\% & 97.1\% & 93.6\% & 95.5\% \\
  SealID & 62.4\% & 79.1\% & 92.8\% & 63.3\% & 79.9\% & 85.6\% & \textbf{98.1\%} & 95.1\% \\
  SeaTurtleIDHeads & \textbf{96.1\%} & 91.2\% & 94.8\% & 82.2\% & 80.3\% & 92.2\% & 95.6\% & 93.9\% \\
  SMALST & 98.0\% & \textbf{100.0\%} & \textbf{100.0\%}$^\dagger$ & 99.9\% & 98.9\% & \textbf{100.0\%} & 99.9\% & \textbf{100.0\%} \\
  StripeSpotter & 93.3\% & 97.6\% & \textbf{98.8\%} & 95.7\% & 81.1\% & 97.0\% & 97.6\% & 97.0\% \\
  WhaleSharkID & \textbf{86.6\%} & 61.6\% & 78.2\% & 34.3\% & 34.6\% & 76.5\% & 51.6\% & 67.4\% \\
  ZindiTurtleRecall & 38.3\% & 74.5\% & 72.6\% & 40.6\% & 30.2\% & 43.3\% & 65.8\% & \textbf{82.1\%} \\
  \midrule
  Mean & \textit{88.1\%} & \textit{87.6\%} & \textit{91.4\%} & \textit{75.7\%} & \textit{79.4\%} & \textit{85.7\%} & \textit{84.0\%} & \textbf{\textit{91.9\%}} \\
  \midrule[\heavyrulewidth]
  \multicolumn{9}{l}{\emph{Contour $\blacktriangle$}} \\
  HappyWhale & \textbf{44.0\%} & 34.5\% & 35.8\% & 11.3\% & 6.6\% & 11.0\% & 33.3\% & 39.7\% \\
  HumpbackWhaleID & 86.2\% & 78.0\% & 89.0\% & 35.6\% & 62.8\% & 56.9\% & \textbf{91.8\%} & 87.3\% \\
  NDD20 & 40.3\% & 39.4\% & 62.0\% & 34.7\% & 12.6\% & 51.8\% & 58.9\% & \textbf{81.2\%} \\
  \midrule
  Mean & \textit{56.8\%} & \textit{50.6\%} & \textit{62.3\%} & \textit{27.2\%} & \textit{27.3\%} & \textit{39.9\%} & \textit{61.3\%} & \textbf{\textit{69.4\%}} \\
  \midrule[\heavyrulewidth]
  \multicolumn{9}{l}{\emph{Face $\circ$}} \\
  CTai & \textbf{92.4\%} & 91.1\% & 92.1\% & 67.4\% & 47.4\% & 78.6\% & 84.5\% & 69.9\% \\
  CZoo & 97.9\% & 99.1\% & \textbf{99.5\%} & 90.8\% & 78.5\% & 94.8\% & 96.7\% & 92.9\% \\
  IPanda50 & 52.0\% & 86.8\% & \textbf{90.3\%} & 75.9\% & 71.3\% & 80.4\% & 83.7\% & 86.9\% \\
  LionData & 13.5\% & 25.2\% & 34.8\% & 20.0\% & 46.5\% & 23.2\% & \textbf{51.0\%} & 40.5\% \\
  MacaqueFaces & 98.9\% & 99.1\% & \textbf{99.3\%} & 91.7\% & 36.8\% & 81.7\% & 98.2\% & 69.3\% \\
  \midrule
  Mean & \textit{70.9\%} & \textit{80.3\%} & \textbf{\textit{83.2\%}} & \textit{69.2\%} & \textit{56.1\%} & \textit{71.7\%} & \textit{82.8\%} & \textit{71.9\%} \\
  \midrule[\heavyrulewidth]
  \multicolumn{9}{l}{\emph{Other $\diamond$}} \\
  BelugaID & 72.0\% & 66.6\% & \textbf{72.5\%} & 33.7\% & 3.8\% & 16.5\% & 65.2\% & 71.6\% \\
  NOAARightWhale & 25.7\% & 40.7\% & 47.0\% & 31.9\% & 19.1\% & 36.6\% & 36.8\% & \textbf{65.0\%} \\
  \midrule
  Mean & \textit{48.9\%} & \textit{53.7\%} & \textit{59.8\%} & \textit{32.8\%} & \textit{11.5\%} & \textit{26.6\%} & \textit{51.0\%} & \textbf{\textit{68.3\%}} \\
  \midrule[\heavyrulewidth]
  Mean, all & \textit{79.2\%} & \textit{80.1\%} & \textbf{\textit{84.8\%}} & \textit{66.6\%} & \textit{65.3\%} & \textit{74.5\%} & \textit{79.2\%} & \textit{84.5\%} \\
  \bottomrule
\end{tabular}

\end{table*}

\Cref{tab:main} reports top-1 accuracy on all 29 datasets.
Our method averages 84.5\%, against 80.1\% for MegaDescriptor and 79.2\% for MiewIDv3, both supervised methods requiring extensive labeling of individual animals. HardSpotter outperforms all tested label-free methods, with FlexGlue coming the closest at 79.2\%. 
Only WildFusion, which also requires training identity labels, is comparable to HardSpotter, outperforming it slightly at 84.8\%.
On the patterned species datasets, HardSpotter has the highest mean, 91.9\%, and the best result on GiraffeZebraID, LeopardID2022, FriesianCattle2017 and ZindiTurtleRecall.
It lags far behind only on WhaleSharkID and NyalaData, where MiewIDv3 leads by 19.2 and 26.9 points.
Notably, on the contour and other datasets the proposed method has the highest average accuracy:  higher by 18.0 points on NOAARightWhale and 19.2 on NDD20.
On the remaining datasets of these groups HardSpotter is within 5\% of the best method.
These results demonstrate the potential of the algorithm to combine contour and natural markings for re-identification even on challenging marine species.

On animal face id datasets HardSpotter averages 71.9\%, significantly worse than WildFusion at 83.2\%, as expected where identity lies in the shape of a face rather than in a marking.

\paragraph{Ablation studies.} In an ablation on SealID and WhaleSharkID, averaged over five seeds, comparing the raw descriptors without the flow lowers top-1 accuracy from 95.1\% to 88.1\% and from 67.4\% to 53.1\%, and linear whitening in place of the flow recovers only part of the loss.
A single chance level for every keypoint in place of the per-keypoint threshold lowers it to 94.6\% and 65.2\%, and counting each match as one instead of by its margin lowers it to 93.3\% and 65.1\%.
Removing the centering, or summing the margins of all matches instead of taking the best one, costs at most 0.8 points.
Replacing the location test by geometric verification lowers it to 93.4\% on SealID, whose body bends, but raises it to 68.4\% on WhaleSharkID, where the identifying feature is the constellation of fine spots.
Among the texture descriptors we compared, HardNet8 on HesAffNet keypoints is the most accurate (both in the supplementary material).

\paragraph{Open set.}
\Cref{tab:openset} applies the open-set decision of \cref{sec:openset} to seven datasets, each under three random open-set splits of the toolkit, in which some individuals appear only among the queries.
Averaged over the seven, the method reaches 87.9\% on the known queries and 91.0\% on the unknown ones, a geometric mean of 89.2\%, with no labels and no tuned threshold.
Rejection costs 6.4 points on the known queries, against 94.3\% top-1 when every query is accepted.
Across the three splits, the seven-dataset mean ranges from 87.8 to 88.3\% on known queries and from 89.2 to 92.9\% on unknown ones.
The accuracy on unknown queries is averaged over far fewer individuals, 5 to 43 per dataset against 48 to 371 known ones, and so varies more between splits (per-dataset spreads in the supplementary material).
Open-set metrics are also highly sensitive to annotation mistakes: an individual recorded in the gallery under another identity turns each of its correctly accepted queries into an error. We found such mistakes in our experiments, and the supplementary material shows an example.

\begin{table}[t]
  \centering
  \caption{Open-set re-identification, mean over three splits (\%). Top-1 accepts every query, and BAKS and BAUS~\cite{adam2025animalclef} are the accuracies on known and unknown queries after rejection.}
  \label{tab:openset}
  \small
  \setlength{\tabcolsep}{3pt}
  \fitcolumn{%
  \begin{tabular}{@{}lrcrrrr@{}}
    \toprule
     & & Known / & & & & Geom. \\
    Dataset & Gallery & unknown & Top-1 & BAKS & BAUS & mean \\
    \midrule
    ATRW & 4,329 & 498 / 569 & 98.2 & 95.2 & 83.6 & 89.1 \\
    Cows2021 & 6,932 & 848 / 890 & 97.2 & 84.0 & 86.8 & 85.4 \\
    FriesianCattle2017 & 720 & 112 / 96 & 98.1 & 98.1 & 82.5 & 90.0 \\
    Giraffes & 1,076 & 175 / 141 & 99.6 & 99.4 & 96.9 & 98.1 \\
    HyenaID2022 & 2,461 & 346 / 319 & 83.5 & 73.6 & 91.6 & 82.1 \\
    SealID & 1,665 & 172 / 243 & 94.5 & 82.0 & 97.8 & 89.6 \\
    SeaTurtleIDHeads & 6,015 & 806 / 761 & 88.8 & 83.3 & 97.8 & 90.2 \\
    \midrule
    \emph{Mean} & & & \emph{94.3} & \emph{87.9} & \emph{91.0} & \emph{89.2} \\
    \bottomrule
  \end{tabular}}
\end{table}

\subsection{Qualitative results}
\label{sec:qualitative}

\begin{figure*}[t]
  \centering
  \includegraphics[width=.9\textwidth]{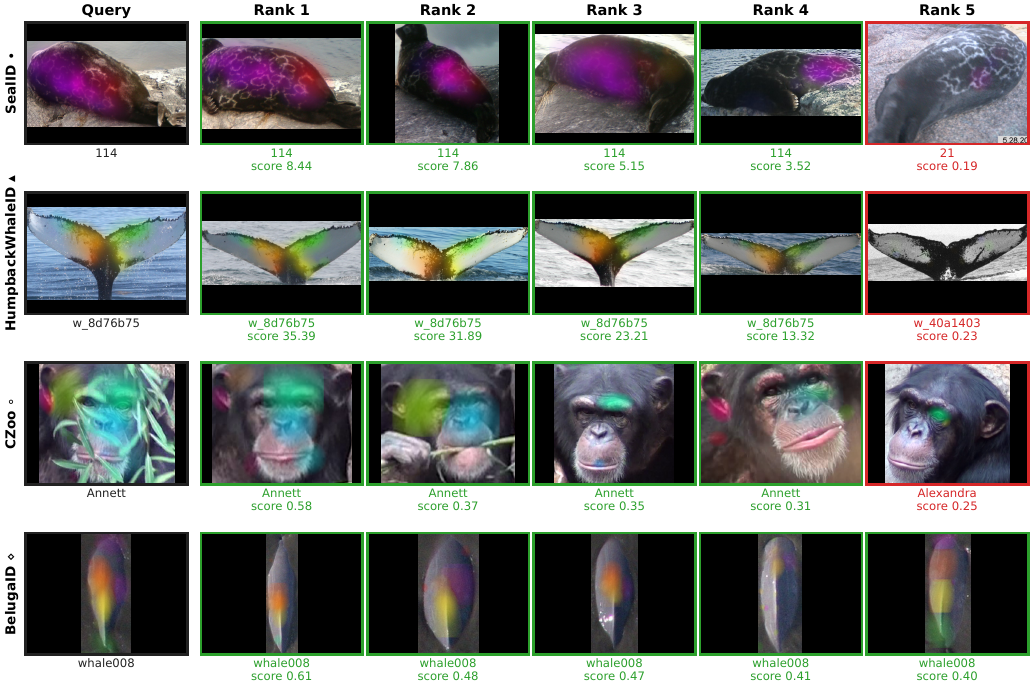}
  \caption{The five best gallery detections for one query from each of four datasets, green for the query's individual and red for others. Glows mark accepted matches, colored by the query keypoint's location token.}
  \label{fig:qualitative}
\end{figure*}

\Cref{fig:qualitative} shows the accepted matches for one query from each of four datasets. For patterned species the method identifies the corresponding body areas and markings, which gives four correct matches in the top five for Saimaa ringed seals. On the seal it matches parts of the rings despite the deformation of the body. On HumpbackWhaleID the four correct matches rest on a combination of natural markings and the shape of the fluke. Although the method is weakest on face re-identification, the CZoo example shows that it still finds identifying features on faces. On beluga whales, one of the hardest cases for human verification, the method picks up features that are difficult to see without a trained eye. The ability to highlight and explain the algorithm's choices helps domain experts verify the outputs of automatic re-identification in real use.

\paragraph{Limitations.}
DINOv2 features confuse the left and right counterparts of a part~\cite{zhang2024telling}, and the location coordinate can inherit this, so a match on the wrong flank can pass the location test.
The scoring pass over keypoint pairs is quadratic in the number of detections.
The method reads DINOv3 only at the keypoints and leaves its image-level vector unused.
On the revisited Oxford and Paris benchmarks (supplementary material), it is ahead of every classical retrieval pipeline on the same keypoints and descriptors, but behind the newer methods that use image-level information, such as AMES and ELViS, and the DINOv3 global token alone is comparable to or ahead of it there.
On the animals the order reverses, and all of them fall far behind ours.
Adding image-level evidence where it helps, without diluting the local count where the global vector is weak, as on seals, is left open.

\section{Conclusion}
\label{sec:conclusion}

Motivated by the need to work without supervision from identity labels, we have proposed a novel animal re-identification algorithm based on matching generic local features and descriptors. Our main technical contribution answers the central question: are two descriptors more similar than chance alone can explain? We do this by applying normalizing flows to map a gallery of descriptors into an approximately normal distribution and derive the answer in closed form. In extensive experiments, the resulting HardSpotter algorithm surpassed all tested unsupervised local feature-based algorithms and all but one supervised algorithm, and it surpassed that one too on species identifiable by body texture or contour patterns. It also accurately determined, without supervision or a manually tuned threshold, when the individual in a query image is not present in the gallery. Every identification rests on keypoint matches that can be drawn on the two images. In short, HardSpotter is a powerful new algorithm for identifying individual animals that is both unsupervised and interpretable.

\iftoggle{cvprfinal}{%
\section*{Acknowledgments}
F.Z. and E.D.M. were funded by by the European Union (ERC, BIOBANG, 101171602). This work was supported by the Imageomics Institute, the ICICLE Institute, and the Finnish Cultural Foundation.
The Imageomics Institute is funded by the U.S. National Science Foundation's Harnessing the Data Revolution (HDR) program under Award No.~2118240, ``Imageomics: A New Frontier of Biological Information Powered by Knowledge-Guided Machine Learning.''
The ICICLE Institute is a U.S. National Science Foundation AI Institute, ``Intelligent Cyberinfrastructure with Computational Learning in the Environment,'' under Award No.~2112606.
We thank the Ohio Supercomputer Center for the compute and storage resources used in this work.
}{}

\section*{LLM Usage Statement}
Claude Opus 5.5 (Anthropic) was used as a coding assistant to implement and run experiments, to assist with the literature review, and to edit parts of the text. The authors checked every result and every statement in the paper and take full responsibility for its content.

\ifsupplonly\else
{
    \small
    \bibliographystyle{ieeenat_fullname}
    \bibliography{main}
}
\fi

\ifsuppl
\clearpage
\ifsupplonly\stopdroppingpages\fi
\maketitlesupplementary

The supplementary material first provides theoretical and practical details of the method in full, in the order of \cref{sec:method}: the descriptors (\cref{sec:suppl-features}), the flow (\cref{sec:suppl-flow}), the per-feature calibration (\cref{sec:suppl-calibration}), the location test (\cref{sec:suppl-location}) and the open-set decision (\cref{sec:suppl-openset-rule}).
It then describes the data and the baselines (\cref{sec:suppl-datasets}) and reports the ablation of the method's parts and the choice of the texture descriptor (\cref{sec:suppl-ablation}), annotation mistakes in the ground truth (\cref{sec:suppl-openset}) and instance retrieval beyond animals (\cref{sec:beyond}).

\section{Descriptors}
\label{sec:suppl-features}

\paragraph{Crop frame.}
After segmenting an animal, a crop is resized. In order to keep the resolution comparable and the aspect ratio preserved, crops are resized to keep the number of DINOv3 tokens about the same, i.e. preserve the area of the detection.
Each detection's box is resized, with its aspect ratio kept, into the closest of 7 preset canvases with aspect ratios from 1:4 to 4:1 and about 1,024 DINOv3 patches of $16 \times 16$ pixels each, $16 \times 64$, $20 \times 51$, $25 \times 40$, $32 \times 32$, $40 \times 25$, $51 \times 20$ and $64 \times 16$ patches, and padded with black.
The mask is transformed in the same way and eroded by $8$ pixels.

\paragraph{Logarithmic map.}
HardNet8 descriptors are unit vectors, $t_i \in \mathbb{S}^{127}$, with 127 degrees of freedom in 128 coordinates.
The logarithmic map sends $\mathbb{S}^{127}$ without the point opposite $\mu$ onto the open ball of radius $\pi$ in its tangent plane at the mean direction $\mu$, and a descriptor $t$ becomes $x = \log_\mu(t) \in \mathbb{R}^{D}$, with $D = 127$.
With $\mu = \bar t / \lVert \bar t \rVert$ the mean direction of the gallery's descriptors, $\theta = \arccos(t \cdot \mu)$, and $Q \in \mathbb{R}^{128 \times 127}$ an orthonormal basis of the tangent plane at $\mu$,
\begin{equation}
  x = \log_\mu(t) = Q^\top \frac{\theta}{\sin\theta} \big(t - (t \cdot \mu)\, \mu\big),
\end{equation}
with $\theta / \sin\theta$ taken as 1 at $\theta = 0$.
The map is exact everywhere except at the point opposite $\mu$, which no descriptor reaches.

\section{The flow}
\label{sec:suppl-flow}

\paragraph{Architecture and training.}
The flow is a stack of $K = 8$ affine coupling layers~\cite{dinh2017density}, each conditioned by a multilayer perceptron with two hidden layers of width 256 and GELU activations, with log-scales bounded to $(-2, 2)$ by $\alpha \mapsto 2\tanh(\alpha/2)$ and the last layer of every conditioner initialized at zero, so that training starts from the identity.
It is trained with Adam at a constant learning rate of $10^{-3}$, batch size 16,384 and weight decay $5 \cdot 10^{-5}$, for 80,000 updates, on the gallery's descriptors with 2\% held out, and the parameters with the best held-out likelihood are kept.

\paragraph{Anchor.}
The likelihood does not uniquely determine the map. Many transformations of the latent, such as rotations or some nonlinear rearrangements, do not affect the likelihood, so multiple maps can minimize the loss.
Independent fits can then end in different maps, which differ in which descriptors come out close.  
We remove this freedom by regularizing the loss with the squared distance between the latent $z = f(x)$ and the linear whitening $C^{-1/2}(x - \bar x)$ of $x$, with $\bar x$ and $C$ the mean and covariance of $x$ over the gallery:
\begin{equation}
  \mathcal{L} = \mathbb{E}\Big[\tfrac12 \lVert z \rVert^2 - \log\lvert \det J_f(x)\rvert\Big]
  + \beta\, \mathbb{E}\Big[\lVert z - C^{-1/2}(x - \bar x) \rVert^2\Big].
  \label{eq:loss}
\end{equation}
Among maps of equal likelihood, the fit then prefers the one closest to the linear solution, and refits land close to each other.
We use $\beta = 0.1$ on every dataset.

\paragraph{Centering.}
Keypoints within one detection share common factors, such as lighting or camera, which should not contribute to re-identification. To remove those effects, we center latents within one detection.
With $\bar z_A$ the mean latent over the keypoints of a detection $A$,
\begin{equation}
  \hat z_i = \frac{z_i - \bar z_A}{\lVert z_i - \bar z_A \rVert} \in \mathbb{S}^{D-1},
  \qquad c(i, j) = \hat z_i \cdot \hat z_j .
\end{equation}

\section{Per-feature calibration}
\label{sec:suppl-calibration}

This section details the per-feature threshold $u_i$ of \cref{eq:ui} in \cref{sec:null}, which rests on three observations.

First, the chance level $u_c$ of \cref{eq:uc} assumes that chance look-alikes are equally dense around every keypoint, while on a fitted flow their density varies, which near the threshold is modeled by one factor $\rho_i$ per query keypoint $i$.
A gallery detection $B$ of $n_B$ keypoints from another animal then holds a match of $i$ at or above a level $u$ with the closed form's probability times $\rho_i$, a model for the small probabilities near the threshold:
\begin{equation}
  \rho_i \big[ 1 - \big(1 - S(u)\big)^{n_B} \big] \approx \rho_i\, n_B\, S(u),
  \label{eq:suppl-density}
\end{equation}
since $n_B\, S(u)$ is small, and the closed form is the case $\rho_i = 1$.
At the threshold of \cref{eq:ui} this probability is
\begin{equation}
  \rho_i\, n_B\, S(u_i) = \frac{\rho_i}{r_i}\, \bar r\, n_B\, S(u_c),
  \label{eq:suppl-rate}
\end{equation}
so once $r_i$ estimates $\rho_i$, every keypoint is accepted by chance at the gallery's average rate $\bar r\, n_B\, S(u_c)$.
On SealID, LeopardID2022 and WhaleSharkID, $\bar r$ is 1.9 to 2.1.

Second, $\rho_i$ is estimated from an order statistic, because a count above a fixed level is often zero for a rare keypoint.
By \cref{eq:suppl-density}, the number of detections of other animals that hold a match of $i$ above $u$ is approximately Poisson with mean $\rho_i L_i(u)$, the $L_i$ of \cref{sec:null} with $u$ in place of $c_{i,(k+1)}$.
At $c_{i,(k+1)}$ this number reaches $k + 1$, so, as for the $(k+1)$-th event of any Poisson process,
\begin{equation}
  \rho_i L_i \sim \mathrm{Gamma}(k + 1, 1), \qquad \mathbb{E}\Big[\frac{L_i}{k + 1}\Big] = \frac{1}{\rho_i}.
  \label{eq:suppl-gamma}
\end{equation}
The estimate $1/r_i = L_i/(k+1)$ is thus unbiased, with a relative standard deviation of $1/\sqrt{k + 1}$ for every keypoint, whatever its $\rho_i$.
This is a $k$-nearest-neighbor density estimate relative to the closed form, with the expected count $L_i$ in place of the volume of the neighborhood.
The numerator is $k + 1$ rather than $k$ because \cref{eq:suppl-rate} divides by $r_i$: the estimate $k/L_i$ is unbiased for $\rho_i$ itself, but would accept every keypoint at $(k+1)/k$ times the average rate.

Third, the order statistic must count chance matches only, so it skips the same-animal candidates of \cref{eq:flag}, the detections that hold far more matches than chance allows.
The candidate test has to run before the candidates are known, so it uses $\bar r_{\mathrm{all}}$, measured over all pairs, which same-animal pairs raise slightly above $\bar r$.

\section{Location test}
\label{sec:suppl-location}

This section gives the exact location agreement of \cref{sec:gate} and shows why the level $g$ of \cref{eq:gate} separates genuine partners from look-alikes.

The tokens are compared in a low-dimensional subspace, which keeps their storage and comparison cheap.
With $\bar s$ the mean of the gallery's tokens and $v_1, v_2, \dots$ the eigenvectors of their covariance, ordered by eigenvalue $\lambda_1 \ge \lambda_2 \ge \dots$, the subspace is
\begin{equation}
  \mathcal{U} = \operatorname{span}\{\bar s, v_1, \dots, v_m\}, \qquad m = \bigg\lceil \frac{(\sum_\ell \lambda_\ell)^2}{\sum_\ell \lambda_\ell^2} \bigg\rceil,
  \label{eq:suppl-subspace}
\end{equation}
where $m$, the participation ratio of the spectrum, is a standard measure of its effective dimension and equals the number of directions when the variance is spread evenly over them.
Each token $s_i$ is projected onto $\mathcal{U}$ as $q_i = P^\top s_i$, where the columns of $P$ form an orthonormal basis of $\mathcal{U}$, and the location agreement is $a(i, j) = q_i \cdot q_j$.
In our experiments, using more dimensions or the full tokens gave the same accuracy.

The level $g$ of \cref{eq:gate} maximizes $S_1 - S_2$, the difference between the shares of rank-one and rank-two matches with location agreement at least $a$.
This also maximizes the share of genuine partners admitted minus the share of look-alikes admitted, because the two ranks differ only in how many genuine partners they contain.
Let $\pi_1$ and $\pi_2$ be the shares of genuine partners among the rank-one and rank-two matches, and $G(a)$ and $F(a)$ the shares of genuine partners and of look-alikes with agreement at least $a$, assumed the same at both ranks.
Then
\begin{equation}
  S_1(a) - S_2(a) = (\pi_1 - \pi_2) \big[ G(a) - F(a) \big].
  \label{eq:suppl-gate}
\end{equation}
As long as texture ranks a genuine partner first more often than second, $\pi_1 > \pi_2$, this factor is positive and does not depend on $a$, so maximizing $S_1 - S_2$ also maximizes $G - F$, without estimating $\pi_1$, $\pi_2$ or $G$.

\section{Open-set decision}
\label{sec:suppl-openset-rule}
This section details the open-set decision of \cref{sec:openset} and gives the closed form of its probability $P_A$.
The decision assumes that the top scores of other animals follow an exponential distribution, so that on a logarithmic scale of rank they lie on a straight line.
It fits this line to the query's scores of other animals and asks whether the best score lies too far above it to be another animal.

Sort the scores of a query $A$ over the gallery as $y_{(1)} \ge y_{(2)} \ge \dots$.
Plotted against the logarithm of their rank $j$, the top scores of other animals then lie on a straight line: every time the rank is multiplied by $e$, the score drops by the same amount $\sigma_A$.
The line should describe other animals, so it is fitted below the level $b_A$ of \cref{sec:openset}, the $(k+1)$-th highest score outside the same-animal candidates.
Every score below $b_A$ enters the fit, including that of a same-animal candidate scoring that low.
Since $m_A$ detections score above it, $b_A$ is the $(m_A + 1)$-th score.

Two neighboring scores $y_{(j)}$ and $y_{(j+1)}$ are one rank apart, which on the logarithmic axis is a step of about $1/j$, so the drop of the line per unit of log rank between them is about $j\,(y_{(j)} - y_{(j+1)})$.
The estimate of $\sigma_A$ averages these drops over the $k$ pairs just below $b_A$,
\begin{equation}
  \hat\sigma_A = \frac{1}{k} \sum_{j = m_A + 1}^{m_A + k} j \,\big( y_{(j)} - y_{(j+1)} \big),
  \label{eq:suppl-slope}
\end{equation}
which under the exponential assumption is the maximum-likelihood estimate.

Extended above $b_A$, the line predicts how many other animals would score as high as the best score.
If each of the $m_A$ scores above $b_A$ came from another animal, it would exceed $b_A$ by an exponential amount of mean $\sigma_A$, and so reach the best score $y_{(1)}$ with probability $e^{-(y_{(1)} - b_A)/\sigma_A}$.
The probability that at least one of them reaches it is
\begin{equation}
  1 - \big(1 - e^{-(y_{(1)} - b_A)/\sigma_A}\big)^{m_A} \approx m_A\, e^{-(y_{(1)} - b_A)/\sigma_A},
  \label{eq:suppl-best}
\end{equation}
which, when small, is the number of other animals the line predicts at the height of the best score.
The query is accepted when this probability is at most $1/\lvert\mathcal{G}\rvert$ (\cref{sec:openset}).

The slope is fitted from only $k$ pairs, so $P_A$ must allow for its error, as Student's $t$-test does when the spread is estimated from a few samples.
With $\tau_A = (y_{(1)} - b_A)/\hat\sigma_A$ the height of the best score above $b_A$ measured in fitted slopes, a height of $\tau_A$ fitted slopes is less surprising than the same height in true slopes, because a slope that came out too small inflates $\tau_A$.
The fitted slope is the mean of $k$ exponential variables, so $\gamma = \hat\sigma_A / \sigma_A$ follows a Gamma distribution with shape $k$ and scale $1/k$, independent of the scores above $b_A$.
Since $(y_{(1)} - b_A)/\sigma_A = \tau_A \gamma$, averaging \cref{eq:suppl-best} over $\gamma$, with $\mathbb{E}_\gamma[e^{-s\gamma}] = (1 + s/k)^{-k}$, gives
\begin{equation}
  \begin{split}
    P_A &= \mathbb{E}_\gamma \Big[ 1 - \big( 1 - e^{-\tau_A \gamma} \big)^{m_A} \Big] \\
        &= \sum_{j=1}^{m_A} (-1)^{j+1} \binom{m_A}{j} \Big(1 + \frac{j\, \tau_A}{k}\Big)^{-k}.
  \end{split}
  \label{eq:openset}
\end{equation}
Since $k$ detections outside the same-animal candidates score above $b_A$, $m_A \ge k$. For $k = 10$ and the smallest $m_A = 10$, a best score seven fitted slopes above $b_A$ gets $P_A \approx 0.044$, about five times the $0.009$ of a known slope.
A query is not accepted when the line cannot be fitted: when $b_A$ or $k$ scores below it do not exist, or when these scores are all equal, as at the zero score of detections without an admitted match.

\section{Datasets, preprocessing and baselines}
\label{sec:suppl-datasets}
\label{sec:suppl-preprocessing}

\paragraph{Preprocessing.}
Each method reads one of five regions of a dataset's images, assigned per dataset in \cref{tab:preprocessing}:
\begin{description}[nosep]
  \item[Whole image] The image as the dataset distributes it, which for several datasets is already a crop of the animal.
  \item[Box] The region of the box the dataset provides.
  \item[Dataset mask] The animal with its background removed by the segmentation the dataset provides, as masks (SMALST), polygons (NDD20) or segmented images with the animal on black (BirdIndividualID, FriesianCattle2015, SealID).
  \item[SAM] The animal segmented by SAM3, prompted with text naming the species, once per prompt, with the results pooled. Masks with a score below $0.3$ are dropped, and duplicates, with a mask IoU of at least $0.2$ or one mask at least 70\% inside another, are removed greedily.
  \item[SAM-center] As SAM, but only the mask that covers the image center is kept, for frames rotated so that the animal of interest lies across the center (FriesianCattle2017).
\end{description}
\paper finds the animal on the full-resolution image and computes its keypoints, descriptors and tokens on the detection's resized crop of about 1,024 patches of $16 \times 16$ pixels (\cref{sec:suppl-features}).

\paragraph{Baselines with identity labels.}
MiewIDv3, MegaDescriptor and WildFusion read the same images: each dataset is prepared once with the WildlifeDatasets tools, and every image is resized so that its shorter side is 518\,px, keeping its aspect ratio.
On these images MegaDescriptor resizes to $384\times384$ and MiewIDv3 to $440\times440$, both without keeping the aspect ratio, in RGB with ImageNet normalization.
WildFusion uses the MegaDescriptor embeddings and runs its local matchers (LoFTR, LightGlue with SuperPoint, DISK or ALIKED) on the same images resized to $512\times512$, LoFTR in grayscale.
WildFusion cannot be calibrated on AAUZebraFish and SMALST (\dag{} in \cref{tab:main}), because in one cross-fitting fold all shortlisted pairs are true matches.

\paragraph{Baselines without identity labels.}
ALFRE-ID~\cite{nepovinnykh2024alfreid}, HotSpotter~\cite{crall2013hotspotter}, LightGlue and FlexGlue read the same 518\,px images as the baselines with identity labels.
ALFRE-ID runs with the background of every image removed by our masks.
HotSpotter runs in the IBEIS system of Wild Me, with each image as one annotation.
LightGlue and FlexGlue score every query against every gallery image, except on the three largest datasets (340 to 430 million query--gallery pairs each), where exhaustive matching is not feasible.
There each query is matched against the 1,000 gallery images closest to it by DINOv3 class-token similarity, which hold a correct gallery image for 89.6\% of the queries on HappyWhale and for more than 99.9\% on the other two.

\begin{table}[t]
  \centering
  \caption{Image preprocessing per dataset, with the regions defined in the text. All baselines read the same 518\,px images, ALFRE-ID with the background outside our masks removed. For SAM and SAM-center, the SAM3 text prompts are in parentheses.}
  \label{tab:preprocessing}
  \footnotesize
  \setlength{\tabcolsep}{4pt}
  \newcommand*{\two}[2]{\begin{tabular}[t]{@{}l@{}}#1\\ #2\end{tabular}}  %
  \begin{tabular}{lll}
    \toprule
    Dataset & Baselines & \paper \\
    \midrule
    AAUZebraFish       & Box          & \two{SAM (zebrafish,}{fish)} \\
    AerialCattle2017   & Whole image  & Whole image \\
    ATRW               & Whole image  & SAM (tiger) \\
    BelugaID           & Box          & Whole image \\
    BirdIndividualID   & Dataset mask & Dataset mask \\
    Cows2021           & Whole image  & Whole image \\
    CTai               & Whole image  & Whole image \\
    CZoo               & Whole image  & SAM (chimpanzee) \\
    FriesianCattle2015 & Dataset mask & Dataset mask \\
    FriesianCattle2017 & Whole image  & \two{SAM-center (cow,}{cattle)} \\
    Giraffes           & Whole image  & SAM (giraffe) \\
    GiraffeZebraID     & Box          & \two{SAM (giraffe,}{zebra)} \\
    HappyWhale         & Whole image  & \two{SAM (whale,}{dolphin)} \\
    HumpbackWhaleID    & Whole image  & \two{SAM (humpback whale,}{whale tail)} \\
    HyenaID2022        & Box          & \two{SAM (hyena,}{animal)} \\
    IPanda50           & Whole image  & Whole image \\
    LeopardID2022      & Box          & SAM (leopard) \\
    LionData           & Whole image  & Whole image \\
    MacaqueFaces       & Whole image  & Whole image \\
    NDD20              & Whole image  & Dataset mask \\
    NOAARightWhale     & Whole image  & \two{SAM (right whale,}{whale)} \\
    NyalaData          & Whole image  & \two{SAM (nyala,}{antelope)} \\
    OpenCows2020       & Whole image  & Whole image \\
    SealID             & Dataset mask & Dataset mask \\
    SeaTurtleIDHeads   & Whole image  & Whole image \\
    SMALST             & Dataset mask & Dataset mask \\
    StripeSpotter      & Box          & Box \\
    WhaleSharkID       & Box          & SAM (whale shark) \\
    ZindiTurtleRecall  & Whole image  & Whole image \\
    \bottomrule
  \end{tabular}
\end{table}

\section{Ablation}
\label{sec:suppl-ablation}
\label{sec:analysis}
\begin{table*}[t]
  \centering
  \caption{Ablation of the parts of the method, top-1 accuracy, mean over five runs with different seeds. Each row removes one part or replaces it by its nearest alternative, with everything else fixed.}
  \label{tab:ablation}
  \small
  \begin{tabular}{lcc}
  \toprule
   & SealID & WhaleSharkID \\
  \midrule
  Full method & \textbf{95.1} & 67.4 \\
  \midrule
  Global threshold in place of the local one & 94.6 & 65.2 \\
  Global threshold tuned on labels in place of the derived one & 94.9 & 65.7 \\
  Raw descriptors, no logarithmic map and no flow & 88.1 & 53.1 \\
  Linear whitening in place of the flow & 92.9 & 61.9 \\
  No centering & 94.2 & 66.9 \\
  Lowe's ratio test in place of the threshold & 88.2 & 49.0 \\
  Location test off & 93.5 & 66.0 \\
  Geometric verification in place of the location test & 93.4 & \textbf{68.4} \\
  Sum pooling in place of the max & 94.9 & 66.9 \\
  Plain count in place of the hinge & 93.3 & 65.1 \\
  \bottomrule
\end{tabular}

\end{table*}

This section gives the full ablation summarized at the end of \cref{sec:main}: the parts of the method in \cref{tab:ablation} and the texture descriptor in \cref{tab:descriptors}.
Each row of \cref{tab:ablation} changes one part of the method on SealID and WhaleSharkID and keeps the rest fixed:
\begin{description}[nosep]
  \item[Global threshold] Every match is held to the chance level $u_c$ of \cref{eq:uc} instead of the threshold $u_i$ of its query keypoint.
  \item[Global threshold tuned on labels] $u_c$ is replaced by one level per dataset, chosen from 0.30 to 0.55 in steps of 0.01 by leave-one-out retrieval on the gallery with the gallery's identity labels.
  \item[Raw descriptors] The centered HardNet8 descriptors are compared directly, with neither the logarithmic map nor the flow.
  \item[Linear whitening] The flow is replaced by the whitened coordinates $C^{-1/2}(x - \bar x)$, the target of the anchor (\cref{sec:suppl-flow}).
  \item[No centering] The latent is normalized without subtracting its detection's mean.
  \item[Lowe's ratio test] A query keypoint keeps its best match in a gallery detection when its distance is below 0.8 times that of the second best in the same detection~\cite{lowe2004distinctive}. The test sets no threshold to measure a margin from, so each kept match scores one.
  \item[Location test off] The indicator of \cref{eq:score} is removed.
  \item[Geometric verification] The matches above the threshold are verified by MAGSAC++~\cite{barath2020magsac}, and a pair is scored by its inliers. The geometric model and its tolerance are chosen per dataset by leave-one-out retrieval on the gallery with the gallery's identity labels, as for the baselines of \cref{tab:beyond}.
  \item[Sum pooling] The best match of a query keypoint is replaced by the sum of its margins over all keypoints of $B$ that exceed its threshold and pass the location test.
  \item[Plain count] Each match scores one instead of its margin.
\end{description}
The rows that change the compared coordinate, raw descriptors, linear whitening and no centering, derive $u_c$, $g$ and the per-keypoint thresholds again in their own coordinate.

The per-keypoint threshold gains 0.5 points on SealID and 2.2 points on WhaleSharkID over the global chance level $u_c$.
A global level tuned with the gallery's labels gains at most 0.5 points over $u_c$, and the full method, which uses no labels, stays 0.2 and 1.7 points above it.

Without the flow, comparing the raw descriptors loses 7.0 points on SealID and 14.3 points on WhaleSharkID.
Linear whitening recovers only part of this loss, 4.8 and 8.8 points, because it applies one metric to the whole sphere, while the flow's Jacobian varies from one descriptor to the next.
Removing the centering costs 0.82 and 0.58 points before rounding, which the table's rounded values show as 0.9 and 0.5 points.

Lowe's ratio test in place of the threshold loses 6.9 points on SealID and 18.4 points on WhaleSharkID.
With no margin to weight by, it is compared with the plain count, which it falls 5.1 and 16.1 points below.

The location test is intended to serve as a more general replacement for geometric verification.
Geometric verification fits one transform between the two photographs, a homography in all our runs, which suits a rigid pattern but not a body that bends between photographs, while the location test only asks that the two keypoints lie at the same place on the body.
The comparison favors geometric verification, whose model and tolerance were chosen with the gallery's identity labels, while the location test uses none.
On SealID, where the seals' bodies bend, the location test gains 1.4 to 1.7 points over no test on every seed, while geometric verification is level with no test.
On WhaleSharkID, whose spot pattern is rigid, geometric verification is ahead of the location test by only 0.3 to 1.3 points, and the location test still gains 1.0 to 1.9 points over no test.

Taking the best match of each query keypoint rather than summing all its matches above the threshold changes little, 0.2 points on SealID and 0.5 points on WhaleSharkID, while scoring each match by its margin instead of as one gains 1.8 and 2.3 points.

\label{sec:suppl-descriptors}
\label{sec:descriptors}
\begin{table*}[t]
  \caption{Choice of the texture descriptor on SealID and WhaleSharkID, top-1 to top-5 accuracy and mAP, mean over three runs with different seeds. In the first block only the descriptor changes, and in the second each descriptor runs with the keypoints its authors use. \dag{} marks runs with detections left out.}
  \label{tab:descriptors}
  \footnotesize
  \setlength{\tabcolsep}{3pt}
  \resizebox{\textwidth}{!}{\begin{tabular}{llcccccccccccc}
  \toprule
   & & \multicolumn{6}{c}{SealID} & \multicolumn{6}{c}{WhaleSharkID} \\
  \cmidrule(lr){3-8}\cmidrule(lr){9-14}
  Descriptor & Keypoints & top-1 & top-2 & top-3 & top-4 & top-5 & mAP & top-1 & top-2 & top-3 & top-4 & top-5 & mAP \\
  \midrule
  \multicolumn{14}{l}{\emph{HesAffNet keypoints, the descriptor exchanged}} \\
  HardNet8~\cite{pultar2020improving} & HesAffNet~\cite{mishkin2018repeatability} & \textbf{95.1} & \textbf{96.7} & \textbf{97.2} & \textbf{97.2} & \textbf{97.4} & 33.4 & \textbf{67.3} & \textbf{71.1} & \textbf{72.6} & \textbf{73.6} & \textbf{74.4} & \textbf{16.7} \\
  HardNet~\cite{mishchuk2017working} & HesAffNet~\cite{mishkin2018repeatability} & 94.7 & 95.7 & 96.5 & 96.6 & 96.6 & \textbf{34.1} & 63.7 & 67.7 & 69.4 & 70.9 & 72.0 & 15.0 \\
  HyNet~\cite{tian2020hynet} & HesAffNet~\cite{mishkin2018repeatability} & 94.2 & 95.2 & 96.2 & 96.3 & 96.6 & 31.7 & 63.2 & 67.3 & 69.1 & 70.8 & 72.2 & 14.8 \\
  SOSNet~\cite{tian2019sosnet} & HesAffNet~\cite{mishkin2018repeatability} & 93.2 & 95.0 & 95.7 & 96.1 & 96.3 & 31.5 & 62.7 & 66.4 & 68.3 & 69.5 & 70.6 & 14.4 \\
  RootSIFT~\cite{arandjelovic2012three} & HesAffNet~\cite{mishkin2018repeatability} & 92.9 & 94.2 & 94.9 & 95.1 & 95.4 & 28.5 & 62.6 & 65.7 & 67.8 & 68.8 & 69.5 & 14.5 \\
  SIFT~\cite{lowe2004distinctive} & HesAffNet~\cite{mishkin2018repeatability} & 92.7 & 94.0 & 94.7 & 95.0 & 95.3 & 28.8 & 62.2 & 65.8 & 67.5 & 68.5 & 69.3 & 14.3 \\
  TFeat~\cite{balntas2016learning} & HesAffNet~\cite{mishkin2018repeatability} & 81.2 & 85.5 & 88.1 & 88.9 & 90.2 & 21.7 & 49.3 & 53.7 & 56.3 & 58.6 & 60.3 & 9.9 \\
  \midrule
  \multicolumn{14}{l}{\emph{Each descriptor with the keypoints its authors use}} \\
  HardNet8 & DoG & 91.0 & 92.9 & 93.7 & 93.9 & 94.0 & 27.0 & 61.7 & 65.5 & 67.3 & 68.3 & 69.2 & 14.1 \\
  SOSNet & DoG & 89.1 & 92.0 & 93.0 & 93.4 & 94.1 & 26.3 & 57.4 & 61.9 & 64.4 & 66.0 & 67.4 & 12.7 \\
  HyNet & Key.Net~\cite{barroso2019keynet} & 88.9 & 90.2 & 91.4 & 91.8 & 92.8 & 22.5 & 60.7\textsuperscript{\dag} & 65.1\textsuperscript{\dag} & 67.1\textsuperscript{\dag} & 68.3\textsuperscript{\dag} & 69.4\textsuperscript{\dag} & 13.8\textsuperscript{\dag} \\
  SIFT & DoG & 88.2 & 90.8 & 91.8 & 92.3 & 92.8 & 22.0 & 57.1 & 61.1 & 63.1 & 63.8 & 64.9 & 12.3 \\
  TFeat & DoG & 76.8 & 80.7 & 82.6 & 84.8 & 85.9 & 18.1 & 46.0 & 50.7 & 53.7 & 55.9 & 57.7 & 8.9 \\
  DISK~\cite{tyszkiewicz2020disk} & DISK & 75.6 & 78.3 & 80.0 & 81.1 & 82.1 & 14.0 & 50.9 & 54.9 & 57.3 & 59.1 & 60.5 & 9.5 \\
  DeDoDe v2~\cite{edstedt2024dedode,edstedt2024dedodev2} & DeDoDe v2 & 70.4 & 78.5 & 80.8 & 82.5 & 84.2 & 16.3 & 42.0\textsuperscript{\dag} & 47.6\textsuperscript{\dag} & 50.5\textsuperscript{\dag} & 52.7\textsuperscript{\dag} & 54.2\textsuperscript{\dag} & 7.7\textsuperscript{\dag} \\
  SuperPoint~\cite{detone2018superpoint} & SuperPoint & 83.0 & 85.1 & 86.4 & 87.5 & 88.0 & 16.6 & 48.1\textsuperscript{\dag} & 52.5\textsuperscript{\dag} & 54.5\textsuperscript{\dag} & 56.1\textsuperscript{\dag} & 57.2\textsuperscript{\dag} & 8.7\textsuperscript{\dag} \\
  ALIKED~\cite{zhao2023aliked} & ALIKED & 86.7 & 88.7 & 90.2 & 91.0 & 91.4 & 19.2 & 60.7\textsuperscript{\dag} & 63.7\textsuperscript{\dag} & 66.0\textsuperscript{\dag} & 66.9\textsuperscript{\dag} & 67.8\textsuperscript{\dag} & 13.6\textsuperscript{\dag} \\
  \bottomrule
\end{tabular}
}
\end{table*}

\Cref{tab:descriptors} compares method performance on different local descriptors.
In the first block every row reads the HesAffNet keypoints and DINOv3 location tokens of all other experiments, and only the descriptor computed on each keypoint's patch changes.
There HardNet8 is the best descriptor at top-1 on both datasets, ahead of HardNet, HyNet, SOSNet and TFeat, which were trained for the same task of patch matching, and of SIFT and RootSIFT.
HardNet and RootSIFT appear only in this block, because their authors evaluated them on Hessian-affine keypoints.
In the second block each descriptor runs with the keypoints and at the image size its authors pair it with, and every one scores below HardNet8 on HesAffNet.
HardNet8 itself loses 4.1 points on SealID and 5.6 points on WhaleSharkID on DoG keypoints, and the learned detector and descriptor systems, SuperPoint, DISK, DeDoDe v2 and ALIKED, stay at least 8.4 and 6.6 points below HardNet8 on HesAffNet.
The choice of HesAffNet and HardNet8 is based on the properties that were enforced during their training, such as affine invariance and strict focus on local texture description with no global context.
SuperPoint's descriptors vary along far fewer directions than their 255 degrees of freedom, so its flow is fitted only on the directions along which they vary enough for the flow to stretch them to unit variance, 95 on SealID and 96 on WhaleSharkID.
Detections with fewer than ten keypoints on the animal are left out (\dag{} in \cref{tab:descriptors}): 7 query and 32 gallery detections for Key.Net, 3 and 3 for DeDoDe v2, 4 and 10 for SuperPoint, and 30 and 132 for ALIKED.

\section{Annotation mistakes in the ground truth}
\label{sec:suppl-openset}

\paragraph{Open-set evaluation.}
Open-set metrics are sensitive to annotation mistakes: when an individual is recorded in the gallery under another identity, each of its queries that the method correctly accepts counts as an error on unknown queries.
\Cref{tab:openset-splits} gives the spread of \cref{tab:openset} over the three splits.
ATRW varies the most, with accuracy on unknown queries from 75.4 to 95.7\%.
On the two splits where it is low, many more queries of absent tigers are accepted, and on inspection at least 8 of them are correct matches to a gallery tiger recorded under another identity (\cref{fig:atrw-labels}), so on these splits the accuracy on unknown queries understates the method.

\begin{figure*}[t]
  \centering
  \includegraphics[width=.6\textwidth]{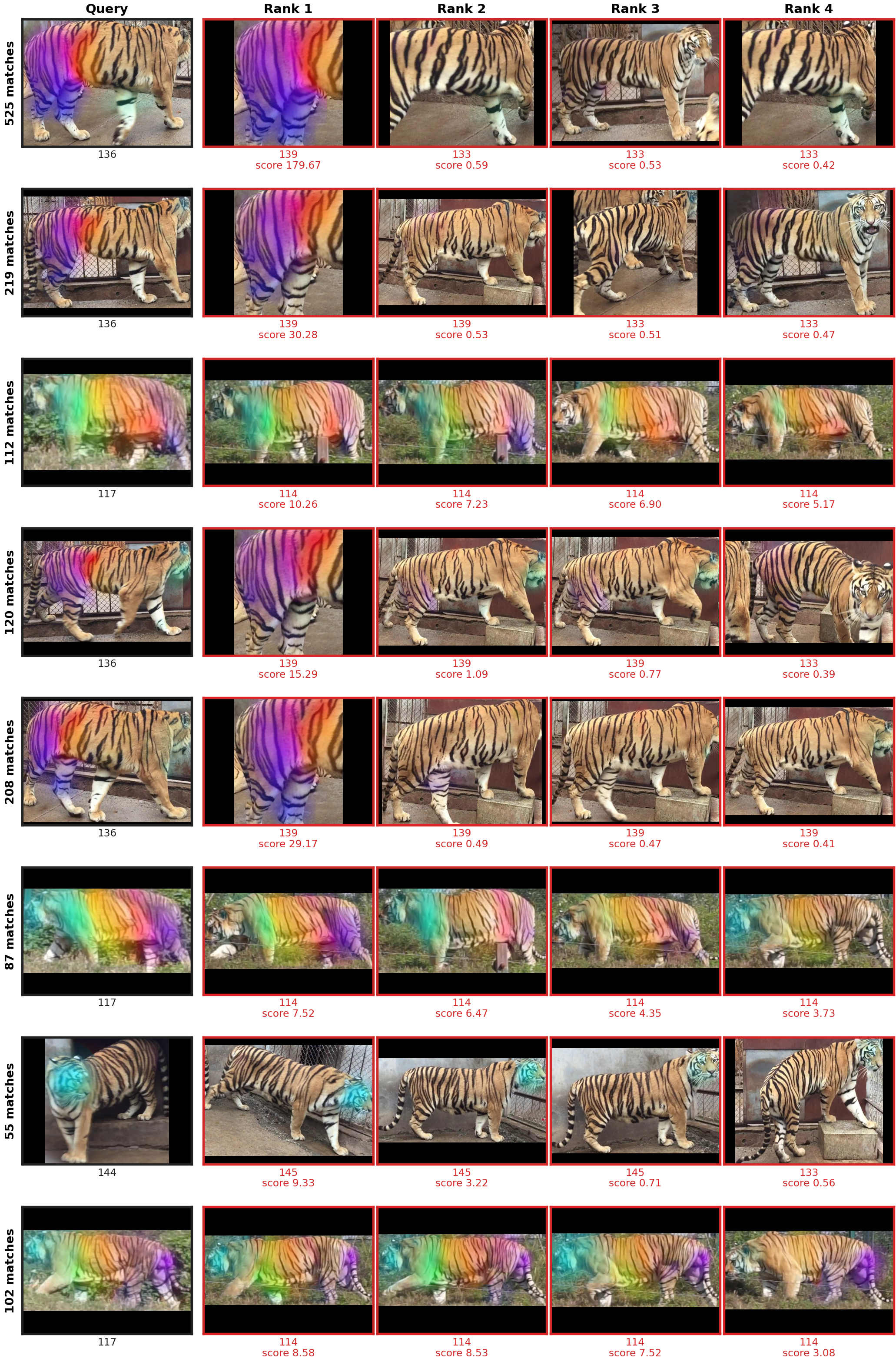}
  \caption{Queries of absent tigers accepted on one ATRW open-set split, with their four best gallery photographs and the accepted matches painted on. Red marks a photograph labeled as another tiger, although each query shows the same tiger as its best match.}
  \label{fig:atrw-labels}
\end{figure*}

\begin{table}[t]
  \centering
  \caption{Open-set re-identification, mean $\pm$ standard deviation over three splits (\%). Gallery and query counts are means over the splits.}
  \label{tab:openset-splits}
  \small
  \setlength{\tabcolsep}{2.5pt}
  \newcommand{\splitsd}[1]{{\scriptsize$\pm$\,#1}}
  \fitcolumn{%
  \begin{tabular}{@{}lrcr@{\,}lr@{\,}lr@{\,}lr@{\,}l@{}}
    \toprule
     & & Known / & & & & & & & \multicolumn{2}{c@{}}{Geom.} \\
    Dataset & Gallery & unknown & \multicolumn{2}{c}{Top-1} & \multicolumn{2}{c}{BAKS} & \multicolumn{2}{c}{BAUS} & \multicolumn{2}{c@{}}{mean} \\
    \midrule
    ATRW & 4,329 & 498 / 569 & 98.2 & \splitsd{0.5} & 95.2 & \splitsd{0.7} & 83.6 & \splitsd{10.7} & 89.1 & \splitsd{5.5} \\
    Cows2021 & 6,932 & 848 / 890 & 97.2 & \splitsd{0.9} & 84.0 & \splitsd{0.8} & 86.8 & \splitsd{3.4} & 85.4 & \splitsd{2.1} \\
    FriesianCattle2017 & 720 & 112 / 96 & 98.1 & \splitsd{1.7} & 98.1 & \splitsd{1.7} & 82.5 & \splitsd{3.8} & 90.0 & \splitsd{1.8} \\
    Giraffes & 1,076 & 175 / 141 & 99.6 & \splitsd{0.7} & 99.4 & \splitsd{0.6} & 96.9 & \splitsd{2.5} & 98.1 & \splitsd{1.2} \\
    HyenaID2022 & 2,461 & 346 / 319 & 83.5 & \splitsd{1.4} & 73.6 & \splitsd{1.1} & 91.6 & \splitsd{5.4} & 82.1 & \splitsd{2.2} \\
    SealID & 1,665 & 172 / 243 & 94.5 & \splitsd{0.4} & 82.0 & \splitsd{1.9} & 97.8 & \splitsd{1.2} & 89.6 & \splitsd{1.5} \\
    SeaTurtleIDHeads & 6,015 & 806 / 761 & 88.8 & \splitsd{1.5} & 83.3 & \splitsd{1.3} & 97.8 & \splitsd{1.0} & 90.2 & \splitsd{0.7} \\
    \midrule
    \emph{Mean} & & & \emph{94.3} & \splitsd{0.6} & \emph{87.9} & \splitsd{0.3} & \emph{91.0} & \splitsd{1.9} & \emph{89.2} & \splitsd{1.1} \\
    \bottomrule
  \end{tabular}}
\end{table}

\paragraph{Closed-set galleries.}
The closed-set galleries hold such mistakes as well, and the score finds them.
For every dataset we took each query's best-scoring gallery detection of a different identity and compared it with the query's best detection of its own recorded identity.
\Cref{fig:label-suspects} shows six such pairs from five datasets, with the score and the rank that every method gives each pair.
In all six, \paper{} ranks the detection of the other identity first, with scores from 33 to 456, while the query's best detection of its recorded identity scores at most 3.3.

On HyenaID2022 and BelugaID the gallery photograph is a near-duplicate of the query, and every baseline ranks the pair first as well.
On WhaleSharkID the gallery photograph is the query's photograph once more, carrying the watermark of the photo-identification library it came from, and only MegaDescriptor does not rank it first.
The other three pairs are not duplicates but different photographs, and there the methods disagree.
On LeopardID2022 the query shows the head and neck and the gallery photograph the whole animal. \paper{} and both learned matchers rank the pair first, while MegaDescriptor and MiewIDv3 rank it 176th and 52nd.
On the first ZindiTurtleRecall pair MiewIDv3 and LightGlue rank the pair first, FlexGlue second and MegaDescriptor 7th.
On the second only \paper{} ranks it first. FlexGlue ranks it third and the two global descriptors 16th and 7th, while LightGlue finds a single match, so this pair rests on our score more than the others do.
The same criterion surfaces near-duplicate pairs on SealID, CTai and MacaqueFaces.

Where every method ranks the pair first, the labels could be right only if all methods failed on the same pair in the same way.
If these pairs are mistakes, each of the queries is counted as an error in \cref{tab:main} against every method that ranks the pair first, so the reported accuracies understate those methods.

\begin{figure}[t]
  \centering
  \includegraphics[width=\linewidth]{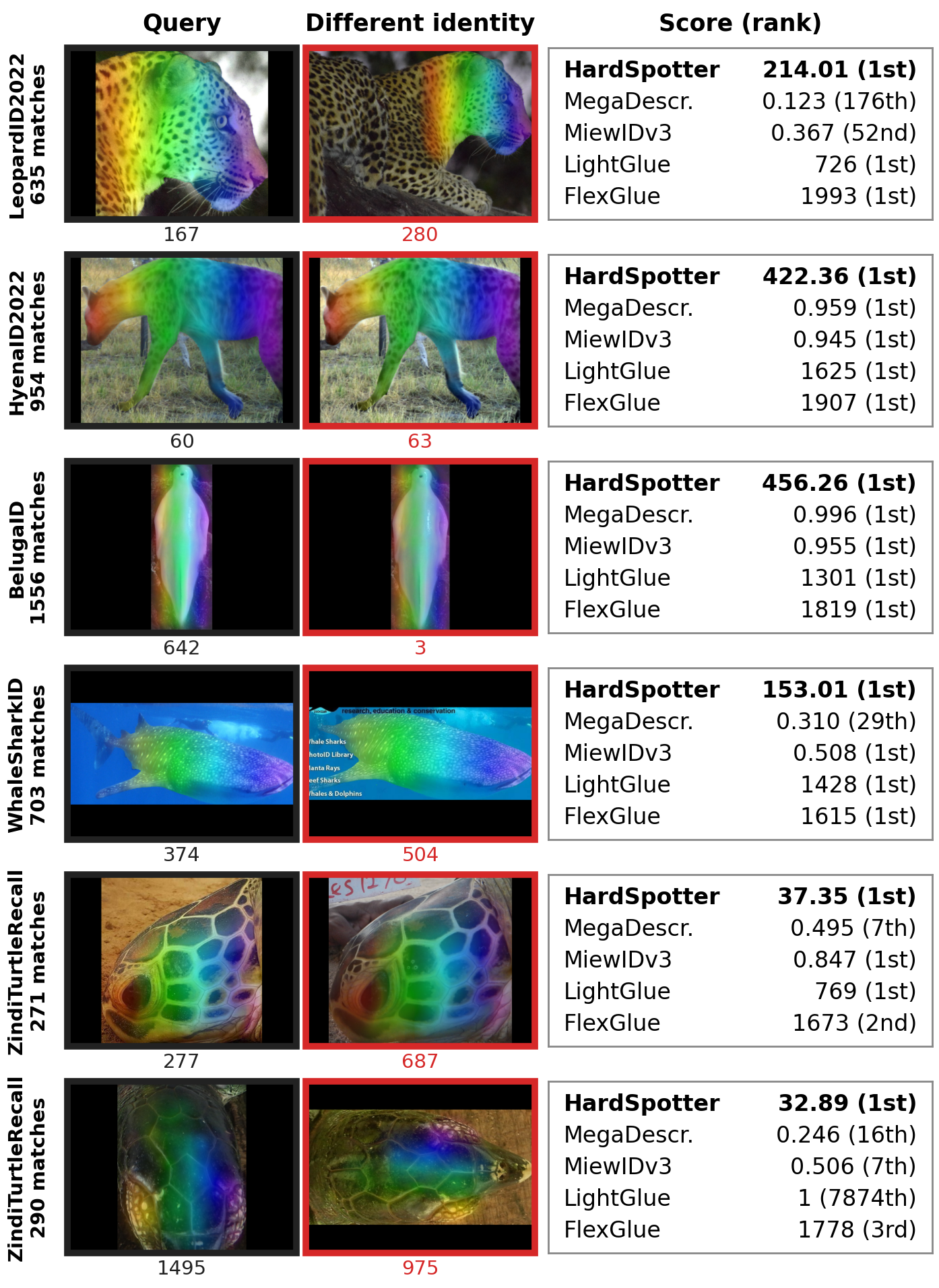}
  \caption{Suspected identity mistakes in closed-set galleries, six pairs from five datasets.
  Each row shows a query (left) and its best-scoring gallery detection recorded under another identity (red frame), with the recorded identities below the images.
  The method's evidence is painted on both images, and the count at the left is the number of matches of the pair.
  The box on the right gives the score of the pair under every method and, in parentheses, the rank of the pair in that method's ranking of the query.}
  \label{fig:label-suspects}
\end{figure}

\section{Instance retrieval}
\label{sec:beyond}
This section gives the instance retrieval results that the limitations in \cref{sec:experiments} refer to.
The method builds on retrieval with local features, so we also apply it, unchanged, to the revisited Oxford and Paris benchmarks~\cite{radenovic2018revisiting}, whose original versions~\cite{philbin2007object,philbin2008lost} served to evaluate HardNet and AffNet~\cite{mishchuk2017working,mishkin2018repeatability}.
We report mAP under the Medium and Hard protocols, with the 70 queries of each benchmark against its whole database and no distractors.
Every row of \cref{tab:beyond} also runs on SealID and LeopardID2022 under the splits of \cref{sec:setup}, 417 queries against 1,663 gallery detections and 1,429 against 5,362.
Our rows are the mean over three runs with different seeds.

\begin{table}[t]
  \centering
  \caption{Instance retrieval on the revisited Oxford and Paris benchmarks~\cite{radenovic2018revisiting}, mAP under the Medium (M) and Hard (H) protocols, and on SealID and LeopardID2022 (Leopard), top-1 and mAP. Our rows and the best result of every column in the first two blocks are in bold, with ties all marked, and + exp.\ marks the expansion of \cref{eq:expansion}.}
  \label{tab:beyond}
  \small
  \setlength{\tabcolsep}{2pt}
  \fitcolumn{%
  \begin{tabular}{@{}lcccccccc@{}}
    \toprule
     & \multicolumn{2}{c}{ROxford} & \multicolumn{2}{c}{RParis} & \multicolumn{2}{c}{SealID} & \multicolumn{2}{c}{Leopard} \\
    \cmidrule(lr){2-3}\cmidrule(lr){4-5}\cmidrule(lr){6-7}\cmidrule(lr){8-9}
     & M & H & M & H & top-1 & mAP & top-1 & mAP \\
    \midrule
    \multicolumn{9}{l}{\emph{HesAffNet + HardNet8}} \\
    Plain sum~\cite{barrow1977parametric} & 39.8 & 19.6 & 51.7 & 25.5 & 55.6 & 11.0 & 65.6 & 11.3 \\
    LNBNN~\cite{mccann2012local} & 57.4 & 35.7 & 55.7 & 29.0 & 89.2 & 29.8 & 85.0 & 26.4 \\
    ASMK*~\cite{tolias2013aggregate} & 56.4 & 32.5 & 60.3 & 32.9 & 82.5 & 20.0 & 80.4 & 20.4 \\
    MAGSAC++~\cite{lowe2004distinctive,barath2020magsac} & 65.2 & 44.0 & 61.1 & 33.5 & 93.3 & 29.4 & 87.2 & 28.5 \\
    \textbf{\paper} & 72.1 & 52.2 & 71.1 & 48.0 & \textbf{95.1} & 33.4 & \textbf{88.3} & 31.8 \\
    HQE-SP~\cite{tolias2014visual} & 67.3 & 44.7 & 69.4 & 45.2 & 80.3 & 22.8 & 79.6 & 24.2 \\
    \textbf{\paper{} + exp.} & \textbf{81.6} & \textbf{58.7} & \textbf{83.7} & \textbf{65.1} & \textbf{95.1} & \textbf{57.8} & \textbf{88.3} & \textbf{48.3} \\
    \midrule
    \multicolumn{9}{l}{\emph{AMES tokens (DINOv2)}} \\
    Plain sum~\cite{barrow1977parametric} & 80.1 & 58.4 & 90.1 & 77.9 & 61.4 & 14.5 & 57.8 & 9.6 \\
    \textbf{\paper} & 80.1 & 58.8 & 76.1 & 54.1 & \textbf{72.8} & 19.7 & \textbf{68.2} & 15.3 \\
    \textbf{\paper{} + exp.} & 87.6 & 69.7 & 85.1 & 70.7 & \textbf{72.8} & \textbf{24.0} & \textbf{68.2} & \textbf{15.6} \\
    ELViS~\cite{suma2026elvis} & 82.6 & 67.3 & 92.5 & 85.0 & 57.6 & 13.6 & 59.6 & 12.6 \\
    AMES~\cite{suma2024ames} & \textbf{92.7} & \textbf{83.7} & \textbf{95.2} & \textbf{90.6} & 58.3 & 14.5 & 63.5 & 13.5 \\
    \midrule
    \multicolumn{9}{l}{\emph{Learned matchers}} \\
    LightGlue (DISK)~\cite{lindenberger2023lightglue} & 79.6 & 58.8 & 68.6 & 46.0 & 85.6 & 15.9 & 85.3 & 25.9 \\
    FlexGlue (ALIKED)~\cite{lindenberger2026flexglue} & 95.2 & 87.7 & 91.2 & 79.7 & 98.1 & 37.9 & 78.6 & 26.5 \\
    \midrule
    \multicolumn{9}{l}{\emph{Global vector}} \\
    DINOv3-L CLS~\cite{simeoni2025dinov3} & 80.0 & 62.2 & 91.5 & 82.0 & 70.7 & 19.7 & 66.1 & 15.9 \\
    \bottomrule
  \end{tabular}}
\end{table}

\paragraph{Baselines.}
\Cref{tab:beyond} is divided into blocks by the features the methods read.
In the first two blocks every row reads the same features, so those rows differ only in how they score them, while the learned matchers read their own features.
Every baseline uses the settings its authors print, and no setting is chosen with the labels of the queries.
Query expansion is used only by HQE-SP and by our rows marked as expanded.

\emph{HesAffNet + HardNet8.}
Our keypoints and descriptors, trained on image patches that come in part from photographs of landmarks~\cite{mishkin2018repeatability,pultar2020improving}.
No row of this block is trained on animals.
\begin{description}[nosep]
  \item[Plain sum] The best cosine of each query descriptor in the candidate, added up in the query direction only, as in chamfer matching~\cite{barrow1977parametric}.
  \item[LNBNN] \cite{mccann2012local}, with the neighborhood depth $k = 10$ of its authors.
  \item[ASMK*] \cite{tolias2013aggregate}, with its codebook trained without labels, on the other city for the city benchmarks and on the dataset's own gallery for the animals.
  \item[MAGSAC++] Exhaustive matching with Lowe's ratio test, verified by MAGSAC++~\cite{barath2020magsac} and scored by its inliers. The ratio and the geometric model are chosen on the other city for the city benchmarks, as is customary, and by leave-one-out retrieval on the gallery for the animals.
  \item[HQE-SP] Hamming query expansion with spatial verification~\cite{tolias2014visual}, one of the two retrieval pipelines in which HardNet was evaluated~\cite{mishchuk2017working}. Its codebook is trained as for ASMK*, and its inlier tolerance is chosen as for MAGSAC++.
\end{description}

\emph{AMES tokens (DINOv2).}
The local descriptors released with AMES and used by ELViS, last-layer DINOv2 tokens~\cite{oquab2024dinov2}, of which AMES keeps those with the highest learned attention.
The attention is trained on Google Landmarks v2~\cite{weyand2020google}, so the selection carries the domain of the city benchmarks, and through self-attention each token carries the context of the whole image.
On the animals the tokens are extracted by the pipeline that produced the AMES release, on each detection with its longest side at 768 pixels and restricted to the animal's mask.
\begin{description}[nosep]
  \item[Plain sum] As in the first block.
  \item[ELViS] \cite{suma2026elvis}, trained on labeled retrieval data. It reports its local score alone, as its authors do, re-ranking the top 400 of the DINOv2 class-token ranking.
  \item[AMES] \cite{suma2024ames}, trained on labeled retrieval data, in its published configuration. It re-ranks the top 1,600 of the SuperGlobal ranking~\cite{shao2023global} on the cities and of the DINOv2 class-token ranking on the animals, the global term with which its published weights were tuned.
\end{description}

\emph{Learned matchers.}
Each reads its own features, ranks every pair by its count of surviving matches as in \cref{tab:main}, and is trained on labeled correspondences.
\begin{description}[nosep]
  \item[LightGlue (DISK)] LightGlue~\cite{lindenberger2023lightglue} on DISK keypoints and descriptors~\cite{tyszkiewicz2020disk}, fine-tuned on MegaDepth, crowd-sourced photographs of landmarks, the domain of the cities.
  \item[FlexGlue (ALIKED)] FlexGlue~\cite{lindenberger2026flexglue} reading ALIKED keypoint locations~\cite{zhao2023aliked} with its own dense features on a DINOv3 ConvNeXt backbone, trained on eight indoor and outdoor datasets that include MegaDepth.
\end{description}

\emph{Global vector.}
The cosine of the DINOv3-L class tokens~\cite{simeoni2025dinov3} of the same canvases our method reads, with no training.

\paragraph{Expansion.}
As HQE-SP expands the query, we also report an expansion of our scores.
A path $A \to B \to C$ from a query $A$ through another detection $B$ to a candidate $C$ says that $A$ and $C$ show the same object only if both of its links hold, so the evidence it carries is its weaker link, and the best path is the widest one.
With $\mathcal{P}$ the pool, the detections scored against one another, which under the protocol of \cref{sec:setup} is the gallery, the expanded score is
\begin{equation}
  \begin{split}
    &\mathrm{score}^{\uparrow}(A, C) = \max\Big( \mathrm{score}(A, C),\\
    &\qquad \max_{B \in \mathcal{P} \setminus \{A, C\}} \min\big( \mathrm{score}(A, B), \mathrm{score}(B, C) \big) \Big),
  \end{split}
  \label{eq:expansion}
\end{equation}
and the candidates are ranked by $\mathrm{score}^{\uparrow}$, with ties broken by $\mathrm{score}$.
Under the minimum, an intermediate can raise a candidate at most to $\mathrm{score}(A, B)$, the query's own evidence for the intermediate, so no threshold is needed.
No path lifts a candidate above the candidate with the highest direct score, which is itself in the pool, and ties are broken by the direct score, so the expansion keeps the top-1 and reorders only the candidates below it, which changes mAP but not top-1 accuracy.
The scores among the pool come from one pass of the gallery against itself, and for each query the expansion is one minimum and one maximum over the rows of the pool's score matrix.

\paragraph{Results.}
On identical features, our method is ahead of every classical pipeline.
Without expansion it is above geometric verification by 6.9 / 8.2 points of mAP on Oxford and 10.0 / 14.5 points on Paris, although buildings are rigid and geometry is at its strongest there, and with expansion it is above HQE-SP by 14.3 / 14.0 and 14.3 / 19.9 points.
It also leads this block on both animal datasets, and of all rows only FlexGlue is ahead, on SealID, at 98.1\% against 95.1\%.
On the cities, the methods that use image-level information are ahead of our method on HardNet8: FlexGlue, AMES~\cite{suma2024ames}, which combines a global and a local score, ELViS~\cite{suma2026elvis}, whose tokens carry the context of the whole image, and a DINOv3 global vector~\cite{simeoni2025dinov3}, which without any training is comparable to our best HardNet8 result on Oxford under the Medium protocol and above it elsewhere.
On the AMES tokens our direct score loses its advantage.
Without expansion, our score is level with a plain sum of best similarities on Oxford and 14.0 / 23.8 points below it on Paris, and ELViS~\cite{suma2026elvis}, a local matcher trained on the same domain, is ahead of ours on both cities.
Expansion raises our score by 7.5 / 10.9 points on Oxford and 9.0 / 16.6 points on Paris, our highest values on the cities, but neither ELViS nor AMES is expanded, so the comparison with them rests on the direct score.
On landmarks much of the identity lies in the image as a whole, the facade and the scene around it, and the descriptors that capture it are those on which our chance level fails, for the reason the next paragraph gives.

\paragraph{Why the chance level needs local descriptors.}
The chance level of \cref{sec:null} assumes that, for two unrelated images, the centered coordinate of a query keypoint is independent of those of the other image's keypoints and spread over all $D$ dimensions.
A patch descriptor such as HardNet8 is computed from the patch alone, the conditions it shares with the rest of its image are removed with the detection mean (\cref{sec:flow}), and two patches of unrelated images are different surfaces, so the assumption holds approximately.
A transformer token is computed from the whole image, since attention mixes every patch into every token, so it carries an image-level part beside its local one.
If that part is kept, it enters every cosine between the two images, and the score of a pair of unrelated images turns in part into a measure of how much the two images resemble each other as a whole.
If it is subtracted, nothing guarantees that the residual spreads over all $D$ dimensions.
The tail of the cosine distribution decays as $(1 - u^2)^{(D-1)/2}$, so with $S_d$ the tail of \cref{eq:S} in $d$ dimensions, a residual spread uniformly over only $d < D$ dimensions raises the tail at $u_c$, for every pair, by roughly
\begin{equation}
  \frac{S_d(u_c)}{S_D(u_c)} \approx \big(1 - u_c^2\big)^{-(D-d)/2},
  \label{eq:dimloss}
\end{equation}
and every lost dimension multiplies the accidental rate by $(1 - u_c^2)^{-1/2}$.
The measurements agree with this account.
Over the pool, matches above $u_c$ occur at 2--4 times the closed form's rate on HardNet8, genuine matches included, and at 15--22 times on the AMES tokens.
On HardNet8 our score gains 19.4 to 39.5 points over a plain sum of best similarities on the same descriptors, and on the AMES tokens it gains 10.4 to 11.4 points in top-1 and 5.2 to 5.7 points in mAP on the animals and nothing or less on the cities.
Leaving the detection mean in place lowers mAP on the cities by up to 4.0 points and changes top-1 on the animals by at most 2.2 points, so neither choice restores what our score gives on HardNet8.

\ifsupplonly
{
    \small
    \bibliographystyle{ieeenat_fullname}
    \bibliography{main}
}
\fi

\fi

\end{document}